\documentclass[onefignum,onetabnum]{siamonline}

\usepackage[utf8]{inputenc}
\usepackage{xspace}
\usepackage{amsfonts}

\def\mathnfa{\mathrm{NFA}}
\def\mathmdl{\mathrm{MDL}}
\usepackage{subcaption}

\ifpdf
\hypersetup{
  pdftitle={The whole and the parts: the MDL principle and the a-contrario framework},
  pdfauthor={R. Grompone von Gioi, I. Ramírez Paulino, G. Randall}
}
\fi
\begin{document}

\title{The whole and the parts: the MDL principle and the a-contrario framework    \thanks{Authors are in alphabetical order. All authors contributed equally to this work.}}
\author{%
    Rafael Grompone von Gioi~%
    \thanks{%
    Universit\'e Paris-Saclay, ENS Paris-Saclay, CNRS, Centre Borelli, France.
    (\email{grompone@ens-paris-saclay.fr})
    } %
    \and %
    Ignacio Ram\'{\i}rez Paulino~%
    \thanks{%
        Instituto de Ingenier\'{\i}a El\'{e}ctrica, %
        Universidad de la Rep\'{u}blica, Uruguay. %
        (%
            \email{nacho@fing.edu.uy}, \email{randall@fing.edu.uy}%
        )%
    } %
    \and %
    Gregory Randall~\footnotemark[3]
}

\maketitle

\begin{abstract}
This work explores the connections between the Minimum Description Length (MDL)
principle as developed by Rissanen, and the a-contrario framework for structure
detection proposed by Desolneux, Moisan and Morel.  The MDL principle focuses
on the best interpretation for the whole data while the a-contrario approach
concentrates on detecting parts of the data with anomalous statistics.
Although framed in different theoretical formalisms, we show that both
methodologies share many common concepts and tools in their machinery and yield
very similar formulations in a number of interesting scenarios ranging from
simple toy examples to practical applications such as polygonal approximation
of curves and line segment detection in images.  We also formulate the
conditions under which both approaches are formally equivalent.
\end{abstract}

%
%
\section{Introduction}

The Minimum Description Length principle (MDL), introduced by Rissanen in
1978~\cite{rissanen78} and further developed
in~\cite{barron98,grunwald07,rissanen84,rissanen86,rissanen92}, is an
information-theoretic approach to the statistical problem of Model Selection.
The MDL principle was developed as a practical, computable approach to the
Algorithmic Information Theory developed by
Solomonoff~\cite{solomonoff1,solomonoff2}, Kolmogorov~\cite{kolmogorov} and
Chaitin~\cite{chaitin1969,chaitin}.  It was later significantly improved by the
development of Universal Coding Theory~\cite{cover06}, a powerful
generalization of the optimal coding methods developed by Shannon, Fano, Elias
and Huffmann.

The a-contrario detection theory was developed by Desolneux, Moisan and
Morel~\cite{DMM2000,DMM_book} and is based on a statistical formulation of
the~\emph{non-accidentalness}
principle~\cite{albert-hoffman1995,wagemans-nonaccidental}.  Its aim is to
control the expected number of false detections under random conditions. Its
rationale is that events likely to arise by accident should not be considered
meaningful detections and must be rejected.  In other words, only significant
deviations from randomness are meaningful.

The theoretical and philosophical foundations of both approaches are very
different.  As the name suggest, the Minimum Description Length principle
follows the same basic idea of Algorithmic Complexity, preferring the model
that leads to the shortest description of the whole data.  The
non-accidentalness principle, on the other hand, suggests rejecting events that
are expected to be observed in random data. The former considers a global
description whereas the latter concentrates on the detection of anomalous
structures present in the data.

Despite their different formulations, both methods find common ground in one
important case: uniformly distributed random data (without loss of generality,
over a finite alphabet).  According to Kolmogorov's definition, the shortest
description of such data will be as long as the data itself with overwhelming
probability. In other words, when the data is random, no clever model can be
devised which allows for a more compact description than spelling out the data
itself. In MDL, such data is non-compressible and thus the best model is the
uniform distribution. In the context of a detection problem this constitutes
a non-detection.  As for a-contrario, the non-accidentalness principle also
guarantees that no detection will occur in this scenario.

This work seeks to show that similarities between both approaches exist beyond
the aforementioned case. We demonstrate this in various examples, ranging from
simple problems, where we can perform an in-depth analysis of the expressions
involved, to more complex real applications where we show, by experimentation,
that both methodologies indeed produce very similar results. In particular, we
develop and evaluate automatic methods to choose the ``best'' polygonal
approximation of a given object, and for detecting line segments, using both
criteria.  Finally, we establish certain conditions under which both approaches
are equivalent.

The MDL principle is mainly a model selection tool, but can also be used as a
detection criterion.  On the other hand, the a-contrario framework was mainly
proposed as a detection criterion, but can also be used for model selection.
Thus, both methodologies can be applied to a large range of similar
applications, facilitating the comparison.  Also, both methodologies require a
modeling step. In all cases, from toy examples to real applications, our
results show that both a-contrario and MDL give very similar results whenever
similar modeling criteria are used.

The rest of this document is organized as follows.  The basic common notation and
concepts are introduced in \Cref{sec:notation}.  Then, \Cref{sec:mdl} provides
a self-contained, concise introduction to the MDL principle, starting with its
Algorithmic Information Theory foundations, and the key tools in Universal
Coding Theory which enable its application in practice. \Cref{sec:nfa} does the
same with the a-contrario theory.  \Cref{sec:square} compares both approaches
in a first toy setting: detecting an individual square on a noisy image.
\Cref{sec:squares} moves to the more interesting case of detecting many
squares.  Next, two more complex problems are addressed.  
\Cref{sec:poly}, deals with the selection of the best polynomial
approximation to a curve; this is a typical model selection problem for which
MDL was specifically developed.  
\Cref{sec:lsd},
deals with  detection of line segments in images; this, in turn, is a
typical computer vision application of the a-contrario approach.  Then,
\Cref{sec:equivalence} establishes the theoretical conditions under which both
approaches are equivalent.  Final comments and perspectives are given in
\Cref{sec:discussion} and \Cref{sec:conclusion}.

\section{Notation, conventions and common concepts}\label{sec:notation}

\def\vx{\mathbf{x}}
\def\mX{\mathbf{X}}

In this brief preamble we summarize the meaning of common terms and symbols
used throughout the document. The object of our study will be a data vector
$\vx=(x_1,x_2,\ldots,x_n)$ of length $n$ where each $x_i$ takes values in a
finite alphabet $\mathcal{X}$ of size $|\mathcal{X}|$; here $|\cdot|$ denotes
the cardinality operator. Images will be treated in the same way by
concatenating their rows into a single vector. Without loss of generality, we
will assume that all images are of size $\sqrt{n}\times\sqrt{n}$.

When the nature of $\vx$ is assumed stochastic, $\mX$ denotes the random vector associated to it, and $X_i$ represents the random variable corresponding to the i-th sample of $\vx$, $x_i$.

All the examples of this work involve analyzing \emph{parts} of the (whole)
data vector $\vx$.  We define a part $\vx_i$ of $\vx$ to be an arbitrary
non-empty subset of elements of $\vx$: $\vx_i=\{x_{j_1},\ldots,x_{j_{n_i}}\}$
where $j_k$ are the indexes of the elements and $n_i$ is the size of the part.
In general, we may analyze several
different parts of the data simultaneously. Let
$\{\mathbf{x}_1,\ldots,\mathbf{x}_N\}$ be a set of $N$ parts of $\vx$ in which
we are interested. This set of parts is  not necessarily a partition: it does
not need to cover the whole $\vx$, and different parts may overlap.  For
example, if $\vx$ is an image, the parts $\{\mathbf{x}_i:i=1.\ldots,N\}$ may
correspond to a group of selected (possibly overlapping) patches. For a particular part, the value of $\vx_i$ will vary for different
realizations of $\vx$. We say that a \emph{configuration} is a particular
realization of a given part. If $\vx$ is stochastic, $\mX_i$ represents the random vector associated to part $\vx_i$.

Parts are the subject of the tests conducted in this work. When dealing with
detection problems, our goal will be to determine whether a part stands out
from the background or not. In that case, we declare a detection. Otherwise, no
detection is declared. In such scenarios, a ``background'' model is assumed.
This can be, for example, a distribution on $\vx$, such as the uniform
distribution over $\mathcal{X}^n$. In general, we associate a model to an
hypothesis. In the case of detections, the background model is associated to
the null hypothesis $H_0$ (i.e., the part is part of the background), and the
non-null hypothesis $H_1$ corresponds to a detection.

In model selection problems, many different models (hypotheses) are proposed
for a given part; the background model (null hypothesis) is always included
among these.  Note that, as formulated, the detection problem is a special case
of a model selection problem where only one non-null hypothesis is formulated.

In any case, the decision on a part is made by computing a score for each model
and selecting the one with the best score. A score is a function of the model,
the part, and its configuration. However, how exactly this score function is
constructed depends on many factors. Most of the technical work in this paper
is devoted to constructing this function.

\section{The Minimum Description Length principle}\label{sec:mdl}

The Model Selection problem can be stated as follows: which is the best model
to describe a given data?  This is a fundamental problem in Statistics and
Science as a whole.  In its most general formulation (any possible model), this
problem is unsolvable~\cite{kolmogorov,li-vitanyi}.  Therefore, the effort has
been focused on selecting the best model from sets of structured candidates.

The Minimum Description Length~\cite{grunwald07} has its philosophical roots in
the famous ``Occam's Razor'' principle, whose modern interpretation can be summarized by saying that, being equally precise, simpler explanations should be preferred over complex ones. To weight different explanations, MDL draws from the theory of
Algorithmic Complexity (also known as Kolmogorov complexity)~\cite{li-vitanyi}.
This theory, developed independently by
Solomonoff~\cite{solomonoff1,solomonoff2}, Kolmogorov~\cite{kolmogorov} and
Chaitin~\cite{chaitin1969,chaitin}, states that the complexity of a data object
is given by the shortest program that can describe it using a Universal Turing
Machine.

\def\model{M}
\def\modelfam{\mathcal{M}}
\def\vx{\ensuremath{\mathbf{x}}\xspace}

In accordance with the unsolvability of the general Model Selection problem, it
has long been shown that finding the shorter program to describe a given data is a
non-computable problem~\cite{non-computable}.  In view of this difficulty, the  MDL principle reduces its search to a set of candidate descriptions that are easy to evaluate: this is usually a family $\modelfam$ of parametric probability models
and a method for \emph{compressing data} using these models.  In short, given
some data, MDL will choose the model $\model \in \modelfam$ which, when fed to the compression algorithm, yields the shortest description for that data.  The choice of the family of models $\modelfam$ for a particular problem is the modeling step.

Given $\modelfam$, the  problem of designing an algorithm
which will produce the shortest possible description of any given data using the
available models in $\modelfam$ is non-trivial.  This is the subject of Universal Coding
Theory (UC)~\cite{cover06};  much of the literature on MDL deals with the
design and implementation of such algorithms.  Whereas the traditional
(pre-Universal) Coding Theory deals with developing compression algorithms for
sources with fully known probability distributions, UC deals with the problem
of encoding data whose distribution is assumed known only in its general form
(e.g., ``a polynomial plus Gaussian noise'' or a ``Markov Process of arbitrary
order'').

More Formally, let $\modelfam = \{\model_{\theta} : \theta \in \Theta \}$ be a
family of probability models defined over $\vx$ indexed by a parameter
$\theta$.  We denote by $L_\modelfam(\vx|\model_\theta)$ the code-length
assigned by model $\model_\theta$ to $\vx$.  To produce a complete description
of $\vx$, $\model_\theta$ needs to be encoded as well; we denote the joint
encoding of $\vx$ and $\model_\theta$ given the family as
$L_\modelfam(\vx,\model_\theta)$ (the family $\modelfam$ is unique and known
both to the encoder and the decoder, so there is no need to describe it when transmitting $\vx$):
\begin{equation}
L_\modelfam(\vx,\model_\theta) = L_\modelfam(\vx|\model_\theta) + L_\modelfam(\model_\theta).
\label{eq:two-parts-coding}
\end{equation}
The two terms in \eqref{eq:two-parts-coding} play the role of a \emph{fitting
term} and a \emph{penalty term}, respectively, in traditional regularization
methods.  It is not necessary to actually encode the data in a binary stream;
only code-lengths matter.  However, the code-length assignments need to be
valid.  A central theorem in Information Theory (see~\cite{cover06})
establishes that any valid code-length assignment needs to satisfy the Kraft
inequality~\cite{cover91}:
$$
   \sum_{\vx \in \mathcal{X}^n} 2^{-L_\modelfam(\vx,\model_\theta)} \leq 1.
$$
The main problem with the original (called ``two-parts'') formulation of
MDL~\cite{rissanen78} is the arbitrariness in separating
$L_\modelfam(\vx,\model_\theta)$ into $L_\modelfam(\vx|\model_\theta)$ and
$L_\modelfam(\model_\theta)$;  as soon as one wants to describe the model
parameter $\theta$, one faces the problem of choosing a model for $\theta$
itself.  

A fundamental development in MDL was the incorporation of tools from
the Universal Coding, which allow for a provably
optimal, \emph{one-part encoding} of $\vx$ directly in terms of the whole
family $\modelfam$: $L_\modelfam(\vx)$. In these schemes, the parameter
$\theta$ is ``marginalized-out''. A fundamental result~\cite{rissanen84} states that $L_\modelfam(\vx)$ can still
be decomposed into two terms (not to be confused with the \emph{two-parts
coding} mentioned earlier):
$$
   L_\modelfam(\vx) = L(\vx|\modelfam) + L(\modelfam).
$$
The term $L(\modelfam)$ is called \emph{Model Complexity} and depends
solely on the size and richness of the model itself: larger model families are
inherently more complex and thus result in a larger, unavoidable overhead, no
matter the encoding in use. The term  $L(\vx|\modelfam)$, called 
\emph{Stochastic Complexity}, depends on the particularities of the data $\vx$
to be encoded. Again, these two terms play the roles of regularization and
fitting terms, even though the division is not explicit.

The proper application of modern MDL requires one to develop an optimal
one-part coding scheme.  In many cases, however, such schemes are impractical
to implement, and one must settle for an imperfect two-parts coding approach.
In such cases (besides complying with the Kraft inequality), the only sanity
check is that the actual code-lengths produced are better than those obtained
by a trivial, raw encoding for the data that one is working with (e.g., if one
wants to encode digital images of $n$ pixels with $8$ bits per pixel, a useful
code should produce code-lengths well below $8n$ bits for most images).

\section{The a-contrario framework}\label{sec:nfa}

The a-contrario framework was introduced by Desolneux, Moisan and
Morel~\cite{DMM2000,DMM_book} as a general way of selecting detection
thresholds while controlling the number of false detections under a background
or null hypothesis $H_0$.  It can be seen as a formalization of the
\emph{non-accidentalness}
principle~\cite{albert-hoffman1995,wagemans-nonaccidental} which states that an
observed structure is meaningful only when the relationship between its
elements is too regular to be the result of an accidental arrangement of
independent elements.  The idea is illustrated informally in the following
passage:
\begin{quote}
\emph{For so many people connected with the Armstrong case to be travelling by
the same train through coincidence was not only unlikely: it was impossible.
It must be not chance, but design.}\\
Agatha Christie, ``Murder on the Orient Express''
\end{quote}
or even in the more concise words of Ian Fleming in ``Goldfinger'': \emph{Once
is happenstance. Twice is coincidence. The third time it's enemy action.} Both
quotes suggest the idea that a large number of coincidences implies a common
cause.  D.~Lowe expressed the same idea more formally in the context of pattern
detection in digital images:
\begin{quote}
\emph{we need to determine the probability that each relation in the image
could have arisen by accident, $P(a)$.  Naturally, the smaller that this value
is, the more likely the relation is to have a causal interpretation.}\\
David Lowe~\cite[p.\,39]{Lowe85}
\end{quote}
The same idea is the basis of hypothesis testing in statistics.

The a-contrario approach aims at detecting parts of the data with anomalous
statistics. The a-contrario formulation requires: (a) a family of events or
parts to be analysed; (b) a function $\xi(\mathbf{x}_i)$  providing the degree
of anomalousness of a data part $\mathbf{x}_i$; (c) a stochastic model $H_0$
for random data. The latter determines the distribution of random data, which
in turn allows to evaluate whether a given event is common or rare.

The function $\xi$ acts on parts $\mathbf{x}_i$ and produces a real number $y_i
= \xi(\mathbf{x}_i)$.  This function introduces an order among all possible
configurations of $\mathbf{x}_i$, determining the sense in which a vector will
be considered anomalous.  A vector with a large enough $y_i$ value will be
considered anomalous.  To give an example, in an image, patches with large
average pixel values may be considered anomalous.  Finally, a stochastic model
$H_0$ is required for background data; a piece of data $\mathbf{x}_i$ will be
considered anomalous when observing the value $y_i = \xi(\mathbf{x}_i)$ or a
larger one is a rare event under $H_0$.  In some settings, there is a natural
stochastic model $H_0$ for unstructured data; we will see some examples later.
In the absence of particular reasons to specify a model $H_0$, we can follow
Laplace's principle of indifference and assume that all possible realizations
of the data vector are equally probable under $H_0$.  
Correspondingly, let $\mathbf{X}_i$
denote the random vector associated with a part $\mathbf{x}_i$ and $Y_i =
\xi(\mathbf{X}_i)$.

We are now ready to introduce the main ideas of the a-contrario approach.  The
formalism is based on a multiple test procedure as used in
statistics~\cite{HT87} and is very similar to the procedure
in~\cite{gordon2007}.  We want a criterion $F$ such that detections are
declared when $F(i,y_i)\leq\varepsilon$ for a fixed value $\varepsilon$.  The
main idea of the a-contrario approach is to design $F$ to control the expected
number of detections under $H_0$; i.e., when $F$ is applied to random variables
$Y_i$.  In such conditions, any detection would be a false detection.  Here, we
will follow the formulation introduced in~\cite{grosjean-moisan}.

\begin{definition}[Grosjean-Moisan~\cite{grosjean-moisan}]\label{def:nfa}
Let $\{Y_1,\ldots,Y_N\}$ be a set of $N$ random variables.  A function
$F(i,y)$ is a NFA (Number of False Alarms) for the random variables $\{Y_i\}$
if
\begin{equation}\label{eq:expected-false-detections}
   \forall \varepsilon > 0,
   \quad
   \mathbb{E}\left[
                     \sum_{i=1}^N \mathbb{1}_{F(i,Y_i) \leq \varepsilon}
             \right] \leq \varepsilon.
\end{equation}
\end{definition}
In words, the condition in \Cref{eq:expected-false-detections} implies that the
expected number of random variables satisfying $F(i,Y_i)\leq\varepsilon$ is
bounded by $\varepsilon$; this condition is equivalent to
\begin{equation}
  \sum_{i=1}^{N}\mathbb{P}\big(F(i,Y_i) \leq \varepsilon\big) \leq \varepsilon.
\end{equation}
A function $F$ satisfying ~\cref{def:nfa} ensures that the average number of
false detections under $H_0$ is less than $\varepsilon$.  Thus, a NFA allows
controlling the global number of false detections by making detections only
when $F(i,y) \leq \varepsilon$ for the observed value $y$.

\begin{proposition}[Grosjean-Moisan~\cite{grosjean-moisan}]\label{prop:nfa}
Let $\{Y_1,\ldots,Y_N\}$ be a set of $N$ random variables and
$\{\eta_1,\ldots,\eta_N\}$ a set of positive real numbers such that
\begin{equation}
   \sum_{i=1}^N \frac{1}{\eta_i} \leq 1.
\end{equation}
Then, the function
\begin{equation}
   F(i,y) = \eta_i \cdot \mathbb{P}[Y_i \geq y]
\end{equation}
is an NFA.
\end{proposition}

The condition $\sum_{i=1}^N \frac{1}{\eta_i} \leq 1$ allows to apply a
different confidence level $\varepsilon/\eta_i$ to each test while still
controlling the average number of false detections by $\varepsilon$.
In short, a detection will be declared in part $\mathbf{x}_i$ if
\begin{equation}
   \mathrm{NFA}_i = \eta_i \cdot \mathbb{P}\big[\xi(\mathbf{X}_i) \geq \xi(\mathbf{x}_i)\big] \leq \varepsilon,
\end{equation}
where $\varepsilon$ is a fixed value indicating the average number of false
detections one is ready to accept when $\mathbf{x}$ is a realization of
$\mathbf{X} \sim H_0$.  In
particular, setting $\eta_i=N$ for all $i$ (which corresponds to the Bonferroni
correction in multiple test settings), assigns the same risk $\varepsilon/N$ to
each test, while keeping the average number of false detections below
$\varepsilon$. In many practical applications the value
$\varepsilon=1$ is adopted.  Indeed, it allows for less than one false
detection per data (for example an image), which is usually quite tolerable.

A few comments are now in place. First, we are following the setting of
\Cref{sec:notation}, so the elements of $\mathbf{x}$ take values in a finite
alphabet $\mathcal{X}$, and thus $\mathbf{x}\in\mathcal{X}^n$.  But the same
a-contrario framework is well-defined with, for example, real valued vectors.

The second comment is about the functions $\xi$.  We mentioned a single
function $\xi$ to be applied on all parts $\{\vx_1,\vx_2,\ldots,\vx_N\}$.
However, the framework can use a different function $\xi_i$ for each part
$\mathbf{x}_i$, leading to variables $y_i = \xi_i(\mathbf{x}_i)$ and random
variables $Y_i = \xi_i(\mathbf{X}_i)$ under $H_0$.  A simple case where this is
useful is when analysing different kind of parts, each requiring a different
evaluation; for example, some parts may be patches of the image, while other
parts may correspond to region boundaries in the image.  Moreover, it is
sometimes useful to compare alternative interpretations for a given data part;
that is, to evaluate two or more kinds of anomaly that may be present in a data
part.  This is how a-contrario can handle Model Selection.  Formally, a part
vector can be duplicated, such that $\mathbf{x}_i$ and $\mathbf{x}_j$
correspond to the same data, but the observed functions $\xi_i$ and $\xi_j$ are
different, resulting in different tests.  For example, the same image patch may
be evaluated as an anomalous bright regions or as an anomalous dark region.
The interpretation which is more anomalous relative to $H_0$, which is
reflected in a smaller NFA value, is the one to be selected.

A last comment concerns the cardinality of the family of parts.  Up to now, we
assumed a finite number of parts $N$.  However, the framework can be extended
to the case of countable infinite number of tests, provided that
$\sum_{i=1}^\infty \frac{1}{\eta_i} \leq 1.$

To summarize, in order to apply the a-contrario detection paradigm, three
ingredients need to be provided: (a) the family of parts (or events or tests)
to be evaluated, together with the risk distribution $\eta_i$; (b) a function
$\xi$ defining an observed quantity;  (c) a probabilistic model for the
background or null hypothesis $H_0$.  Of course, each of these three components
needs to be worked out for a particular problem.  The choice of these three
components is a modeling step.

\section{Detecting a square on a noisy image}\label{sec:square}

Our first experiment is purposely simple and will serve two objectives: first,
to introduce the technical aspects of both MDL and a-contrario methodologies;
second, to provide a setting that is simple enough to be compared analytically
and where intuition can be easily developed.

\begin{figure}[t]
\centering
\includegraphics[width=0.75\textwidth]{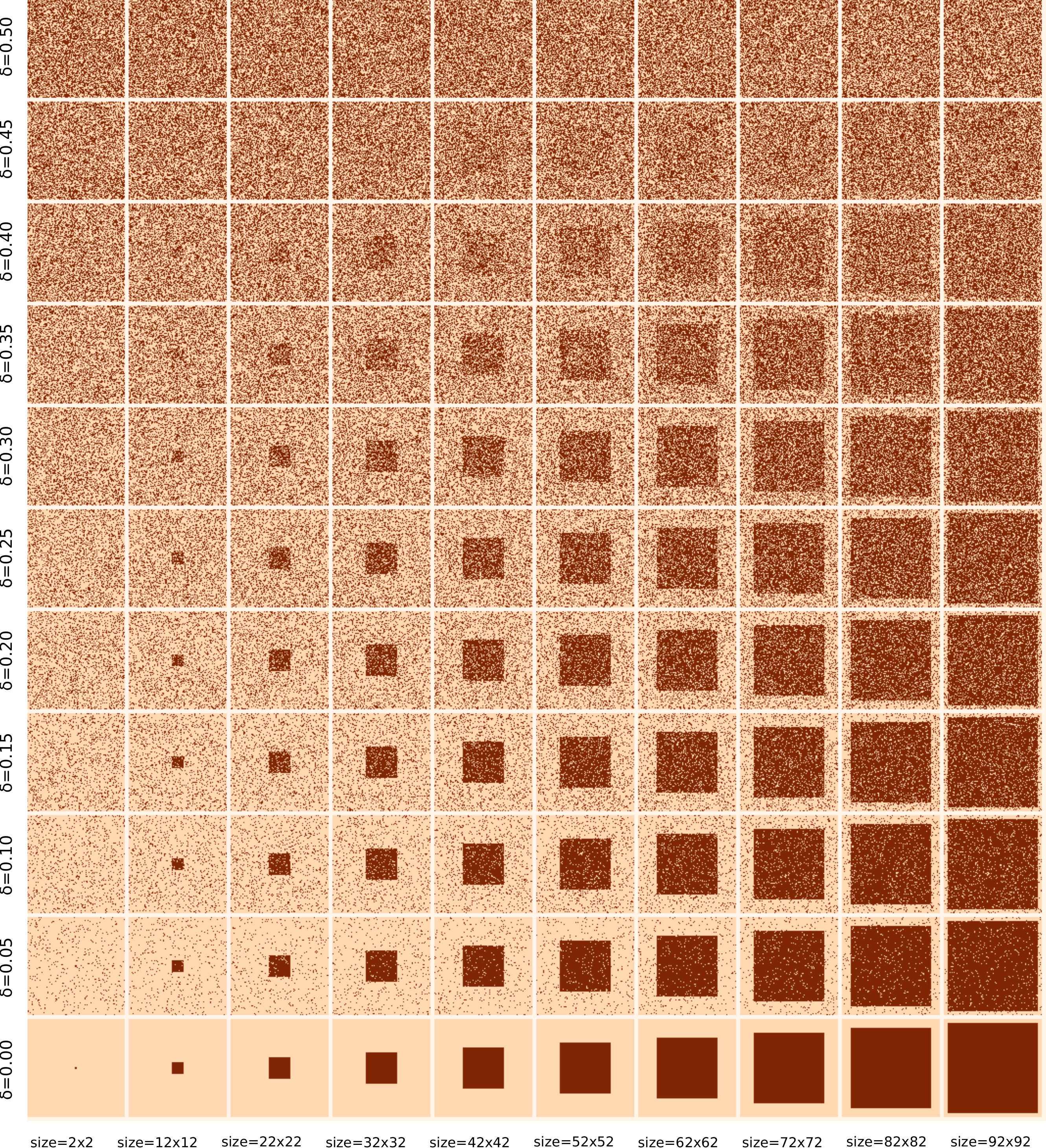}
\caption{Images used in the single square setting.  Noise level increases
upward, whereas square size increases from left to right. At a glance, one can
clearly see the squares in the center and lower images of the grid.
The smaller the square, the faster it vanishes upward.  This also happens when
the square is too large, as the rim  is  confounded with the
square.}
\label{fig:single-square}
\end{figure}

Our object of study is a square image $\vx$ of size $\sqrt{n}{\times}\sqrt{n}$
with  $n$ pixels. The image $\vx$ is binary: its elements $x_{ij} \in
\{0,1\}$ for all $i,j$.  The image is subject to noise, in the sense that the
observed pixels of $\vx$ are the result of an underlying, unobservable image
whose pixels are flipped independently with an unknown probability $0 < \delta
< 1$.  \Cref{fig:single-square} illustrates the setting.

The task is to determine whether a square is present in the underlying
unobserved image based on the noisy observation $\vx$.  Cast as a detection
problem we have two hypothesis: if there is no square, the underlying image
pixels are all $0$'s  and the observed $1$'s are due to some of these $0$'s
being flipped into $1$'s by noise; if a square of size
$\sqrt{n_1}{\times}\sqrt{n_1}$ is indeed present, its unobserved pixels will
have a value of $1$.  If present, the size $n_1$ of the square as well as its
position on the image are also unknown.

If the square is sufficiently large,  we expect roughly $n_1(1-\delta)$ pixels
to be $1$, and   $n_1\delta$ pixels to be $0$.  Correspondingly, the $n-n_1$
background pixels should consist of roughly $(n-n_1)(1-\delta)$ $0$s and
$(n-n_1)\delta$ $1$s. If $\delta < 0.5$, these two quantities should be
distinguishable.  Actually, if $\delta > 0.5$ the same thing would happen, but
the background would be darker than the foreground.  Therefore, w.l.o.g., we
consider error probabilities $0 < \delta < 0.5$ hereafter.

Note that a complete computer vision application usually includes two distinct
stages: the first one produces a set of feasible \emph{candidates} using some
heuristic, and the second one validates the candidates, either jointly or
separately, using some significance criterion. As we are interested in the
latter problem, in this and all the following experiments, we assume the set of
candidates to be fixed and given to us.

We will now work out the aforementioned detection problem using both
frameworks.

\subsection{MDL detection of a noisy square}\label{seq:mdl-square}

In order to formulate the square detection problem under the MDL framework we
consider the task of encoding the binary image $\vx$ under two different
hypothesis: the null hypothesis $H_0$, and the non-null hypothesis $H_1$. Each
will result in a corresponding theoretical code-length: $L_0$ or $L_1$. The
task now is to compute these values for the given image $\vx$.

Under the null hypothesis, the unobserved image is all $0$s and the observed
image is the result of some of these $0$s being flipped to $1$ independently
and with unknown probability $\delta$. The observed image $\vx$ is therefore an i.i.d. Bernoulli process  $\mX$ where $\mathbb{P}(X_i=1)=\delta$.  One possible
universal code for this case is the so-called Enumerative
Coding~\cite{cover73}.  This is a two parts code where the first part uses
$\log n$ bits (if not specified, logarithms are in base $2$) to describe the number $k$ of ones in $\vx$, and the second part
gives the position of $\vx$ within the lexicographical list of all sequences of
length $n$ with $k$ ones; this requires $\log {n \choose k}$ bits:
\begin{equation}\label{eq:square-l0}
  L_0 = \log n + \log {n \choose k}.
\end{equation}

In order to compute $L_1$, we assume that the unobserved image contains a
single square of size $n_1=\sqrt{n_1}\times\sqrt{n_1}$ 1s whose upper-left
corner is located at some arbitrary position $(i,j)$. The observed image $\vx$
is the result of flipping those unobserved pixels independently with unknown
probability $\delta$. Let $k_1$ be the number of 1s inside the square,
$k_0=k-k_1$ the number of 1s outside the square, and  $n_0=n-n_1$ the number of background pixels.  Given this information, we treat both, background and
square, as two independent Bernoulli sequences: the background  with parameter
$q_0=k_0/n_0$, and the square with parameter $q_1=k_1/n_1$. The resulting
code-length is  the concatenation of two codes, both analogous to
$L_0$, plus  $\log n$ bits for describing the location of the square, and
$0.5\log n$ more bits to describe the square side: 
\begin{equation}\label{eq:mdl-square-l1}
  L_1 = \frac{3}{2}\log n + \log n_0 + \log {n_0 \choose k_0}
      + \log n_1  + \log {n_1 \choose k_1}.
\end{equation}

The hypothesis $H_1$ is selected when $L_1 < L_0$. We can express
this condition by defining the \emph{MDL score}, $\mathmdl = L_1 -
L_0$,
\begin{equation}\label{eq:mdl-single-square}
  \mathmdl = \frac{3}{2}\log n + \log n_0 + \log {n_0 \choose k_0}
                + \log n_1  + \log {n_1 \choose k_1} - \log n
                - \log {n \choose k}.
\end{equation}
A positive detection is declared when $\mathmdl < 0$. As a final note, we
remind the reader that this is not the only possible MDL formulation for this
problem; just a reasonable one.

\subsection{A-contrario detection of a noisy square}

In the a-contrario framework, a test is defined for each possible structure to
be detected or part evaluated; in our case each possible square in the image is
a part and defines a test.  Then, a random variable is associated to each test
and an NFA is defined according to \cref{def:nfa}.  Using \cref{prop:nfa},
$\mathnfa = N\,\mathbb{P}(\omega)$, where $N$ is the number of tests and
$\mathbb{P}(\omega)$ is the probability of observing the event $\omega$ under
the null hypothesis (usually called the \emph{background model}).  A detection
is declared when the NFA value for a given event $\omega$ is below a certain
threshold $\varepsilon$; the a-contrario setting ensures that the average
number of false detections under $H_0$ is controlled by $\varepsilon$.

As for the number of tests $N$, there are two unknown quantities that determine
each candidate structure in the current problem: the size of the square, $n_1$,
and its location within the image, which can be specified by the linear index
of the upper-left corner. For an  $\sqrt{n}\times\sqrt{n}$ image we have $n$
possible positions and $\sqrt{n}$ possible values for $n_1$.  This gives
$N=n^{3/2}$ tests.

We assume the same null hypothesis $H_0$ as in the MDL formulation described
before, i.e., the pixel values are independent Bernoulli random variables with
probability $\delta$.  The parameter $\delta$ is unknown, but can be estimated
as the empirical density of $1$s in the image. Likewise, as before, we let $k$
be the total number of $1$s in the image,  $n_1$ be the number of pixels inside the square, and $k_1$ be the number of 1s inside the square.  Following the
same notation, we set $q = k/n$, the fraction of 1s in the image.

In the notation of \Cref{sec:nfa}, each possible square in the image is a part
$\mathbf{x}_i$ and the function $\xi$ count the number of pixels with value one
in the square.  According to the a-contrario approach, the region will be
declared a detection if observing \emph{at least} $k_1$ ones within the square
is highly unlikely under the null hypothesis.  We define the random variable
$K_1$ corresponding to the number of 1s in the square under the null
hypothesis $H_0$.  We are interested in the event $\omega = \{K_1:K_1 \geq
k_1\}$.  Given the empirical density estimation $q$, the probability of this
event is,
$$
   \mathbb{P}[K_1 \geq k_1] = \sum_{i=k_1}^{n_1}{n_1 \choose i} q^i (1-q)^{n_1-i},
$$
which is the tail of a Binomial distribution of parameters $n_1$ and $q$.  We
have now completed our calculation of the NFA:
\begin{equation}\label{eq:nfa-single-square}
    \mathnfa = n^{3/2}\sum_{i=k_1}^{n_1}{n_1 \choose i} q^i (1-q)^{n_1-i}.
\end{equation}
A square is detected when $\mathnfa \leq \varepsilon$. Again, this is not the only possible a-contrario formulation of this problem,
but a reasonable one.  Also, the modelling is similar to the previous MDL
formulation for the same problem.

\subsection{Comparison on a noisy square}\label{sec:mdlvsnfa}

At first sight, the criteria~\cref{eq:mdl-single-square}
and~\cref{eq:nfa-single-square} seem quite unrelated.  We will now develop
further on these expressions and search for common ground from an analytical
point of view.

We begin with \cref{eq:nfa-single-square}, which is somewhat easier.  The
probability term above can be bounded from above using Hoeffding's
inequality~\cite{Hoeffding},
\begin{equation}
  \mathbb{P}[K_1 \geq k_1] \leq \left( \frac{q}{q_1}\right)^{n_1q_1}
                              \left(\frac{1-q}{1-q_1}\right)^{n_1(1-q_1)}
  \quad  \text{for}
  \quad  q_1 > q.
\end{equation}
(When $q_1\leq q$, it is easy to show that $\mathbb{P}[K_1 \geq k_1] \geq
\frac{1}{2}$ and it is not an interesting case.)  We will use this upper-bound
as an approximation to the probability term.  Plugging this
into~\cref{eq:nfa-single-square} and expressing the result in logarithmic
terms, 
 we obtain:
\begin{equation}
  \log \mathnfa \approx \frac{3}{2} \log n + n_1q_1 \log \frac{q}{q_1}
                      + n_1(1-q_1) \log \frac{1-q}{1-q_1}.
\end{equation}
Taking $n_1$ as a common multiplier of the last two terms we get:
\begin{equation}\label{eq:nfa-single-square-2}
\begin{split}
  \log \mathnfa & \approx  \frac{3}{2} \log n
  + n_1 \left[ q_1 \log \frac{q}{q_1} + (1-q_1) \log \frac{1-q}{1-q_1} \right].\\
  & =  \frac{3}{2}\log n - n_1 D(q_1||q),
\end{split}
\end{equation}
where $D(q_1||q)$ is the Kullback-Leibler divergence (KLD) of a Bernoulli
distribution of parameter $q$ with respect to $q_1$.  As expected, the more
dissimilar $q_1$ is to $q$, the smaller the above term will be, and the more
meaningful will be the square detection.  If we choose the usual NFA parameter
$\varepsilon=1$ we get a detection whenever
\begin{equation}\label{eq:nfa-single-square-criterion}
  \log {\mathnfa} \approx \frac{3}{2}\log n - n_1 D(q_1||q) < 0.
\end{equation}
As the KLD is non-negative, the resulting score in \cref{eq:nfa-single-square}
is guaranteed to result in a detection when $q_1$ is sufficiently
distinguishable from $q$, the background model.\footnote{The relation between
the NFA and the Kullback-Leibler divergence was stated in \cite{DMM_book} for
the case of detection of modes of histograms.}

We will now work on \cref{eq:mdl-single-square}.  Using Stirling's
approximation for $\log{n \choose k}$ (see \cref{sec:stirling}) and rearranging
terms we get,
\begin{equation}\label{eq:mdl-square-l0-approx}
\begin{split}
  L_0      &\approx \log n - \frac{1}{2} \log2\pi + \frac{1}{2} \log\frac{n}{k(n-k)}
           + k \log\frac{n}{k} + (n-k) \log\frac{n}{n-k} \\
           &= \log n - \frac{1}{2} \log2\pi + \frac{1}{2} \log\frac{n}{k(n-k)}
           + n \left[ -\frac{k}{n} \log\frac{k}{n} - \frac{n-k}{n} \log\frac{n-k}{n} \right] \\
           &= \log n - \frac{1}{2} \log2\pi + \frac{1}{2} \log\frac{n}{k(n-k)}
           + n\,h(q),
\end{split}
\end{equation}
where $h(q) = -q\log{q} - (1-q) \log (1-q)$ is the Binary Entropy
function~\cite{cover06}.  Similarly,
\begin{equation}\label{eq:mdl-square-l1-approx}
\begin{split}
  L_1      \approx \frac{3}{2} \log n  &+ \log n_0 - \frac{1}{2} \log2\pi
                                       + \frac{1}{2}\log \frac{n_0}{k_0(n_0-k_0)}
                                       + n_0h(q_0) \\
                                       &+ \log n_1 - \frac{1}{2} \log2\pi
                                       + \frac{1}{2}\log \frac{n_1}{k_1(n_1-k_1)}
                                       + n_1h(q_1).
\end{split}
\end{equation}
Combining \cref{eq:mdl-square-l0-approx} and \cref{eq:mdl-square-l1-approx} we
get the corresponding approximate expression for $\mathmdl$:
\begin{equation}\label{eq:mdl-single-square-approx-1}
\begin{split}
  \mathmdl \approx& \frac{3}{2}\log n  + \log n_0 + \log n_1 - \log n
                + n_0h(q_0) + n_1h(q_1) - n\,h(q) \\
                +& \frac{1}{2} \log \frac{n_0}{k_0(n_0-k_0)}
                + \frac{1}{2} \log \frac{n_1}{k_1(n_1-k_1)}
                - \frac{1}{2}\log \frac{n}{k(n-k)}
                - \frac{1}{2} \log2\pi \\
                =& \frac{3}{2}\log n
                + n_0 h(q_0) + n_1 h(q_1) - n\,h(q) \\
                +& \frac{1}{2}\Big[g(k_0,n_0) + g(k_1,n_1) - g(k,n)\Big]
                - \frac{1}{2} \log2\pi.
\end{split}
\end{equation}
where $g(k,n)=\log\frac{n^3}{k(n-k)}$.  Observing that $q = \frac{n_0}{n} q_0 +
\frac{n_1}{n} q_1$, $n = n_0 + n_1$ and $1 = \frac{n_0}{n} + \frac{n_1}{n}$
allows us to write
\begin{equation}\label{eq:h0-split}
\begin{split}
  h(q) &= h\Big( \frac{n_0}{n} q_0 + \frac{n_1}{n} q_1 \Big) \\
       &= - \left[ \frac{n_0}{n} q_0 + \frac{n_1}{n} q_1 \right] \log q
          - \left[ 1 - \frac{n_0}{n} q_0 - \frac{n_1}{n} q_1 \right] \log (1-q) \\
       &=   \frac{n_0}{n} \Big[ -q_0 \log q - (1-q_0)\log (1-q) \Big]
          + \frac{n_1}{n} \Big[ -q_1 \log q - (1-q_1)\log (1-q) \Big],
\end{split}
\end{equation}
and
\begin{equation}\label{eq:h0-split-2}
\begin{split}
n_0 h(q_0) + n_1 h(q_1) - n\,h(q) =& n_0 \Big[ h(q_0) + q_0 \log q + (1-q_0)\log (1-q) \Big] \\
+& n_1 \Big[ h(q_1) + q_1 \log q + (1-q_1)\log (1-q) \Big] \\
=& - n_0 D(q_0||q) - n_1 D(q_1||q).
\end{split}
\end{equation}
Which leads us to:
\begin{equation}
\begin{split}
   \mathmdl
     &\approx
   \frac{3}{2}\log n - n_0D(q_0||q) - n_1 D(q_1||q) \\
      &+ \frac{1}{2}\Big[g(k_0,n_0) + g(k_1,n_1) - g(k,n)\Big]
       - \frac{1}{2} \log2\pi.
\label{eq:mdl-single-square-approx-2}
\end{split}
\end{equation}
As with the a-contrario method, \cref{eq:mdl-single-square-approx-2} is defined
as long as $n,n_0,n_1,k_0,k_1$ are all strictly positive.  Let us repeat
\cref{eq:nfa-single-square-criterion} here for reference:
$$
   \log {\mathnfa} \approx \frac{3}{2}\log n - n_1 D(q_1||q).
$$
As can be seen, both terms in the a-contrario expression are also present in
the MDL one.  The first one embodies the uncertainty in the parameters of the
problem (the size of the image and the square).  The second one measures the
discrepancy between the empirical distribution of 1's in the whole image (the
null hypothesis) and the distribution of 1's inside the square (the non-null
hypothesis).  The MDL expression adds a second KLD term which takes into
account the difference between the distribution of 1's in the image and its
distribution \emph{outside} the square; this reflects the fact that MDL takes
both elements (object and background) into account when making a decision,
whereas a-contrario concentrates on the object to be detected.  Next, MDL
includes a constant term $-(1/2)\log 2\pi$, (which can be disregarded for
sufficiently large $n$). There is one last term which deserves some attention:
\begin{equation}
   \frac{1}{2}\Big[g(k_0,n_0) + g(k_1,n_1) - g(k,n)\Big].
\label{eq:mdl-devil-term-1}
\end{equation}
As shown in \cref{sec:g-bounds}, \cref{eq:mdl-devil-term-1} this term can be
bounded as:
\begin{equation}
    \log n
    \leq
    \frac{1}{2}\Big[g(k_0,n_0) + g(k_1,n_1) - g(k,n)\Big]
    \leq
    \log \frac{n^{\frac{5}{2}}}{4(n-2)} - 1
    \approx
    \frac{3}{2}\log n - 3.
\end{equation}
In words, the above term accounts for an additional difference between MDL and
a-contrario with a magnitude in the order of $\frac{3}{2} \log n$.

\begin{figure}[t]
\centering
\includegraphics[width=\textwidth]{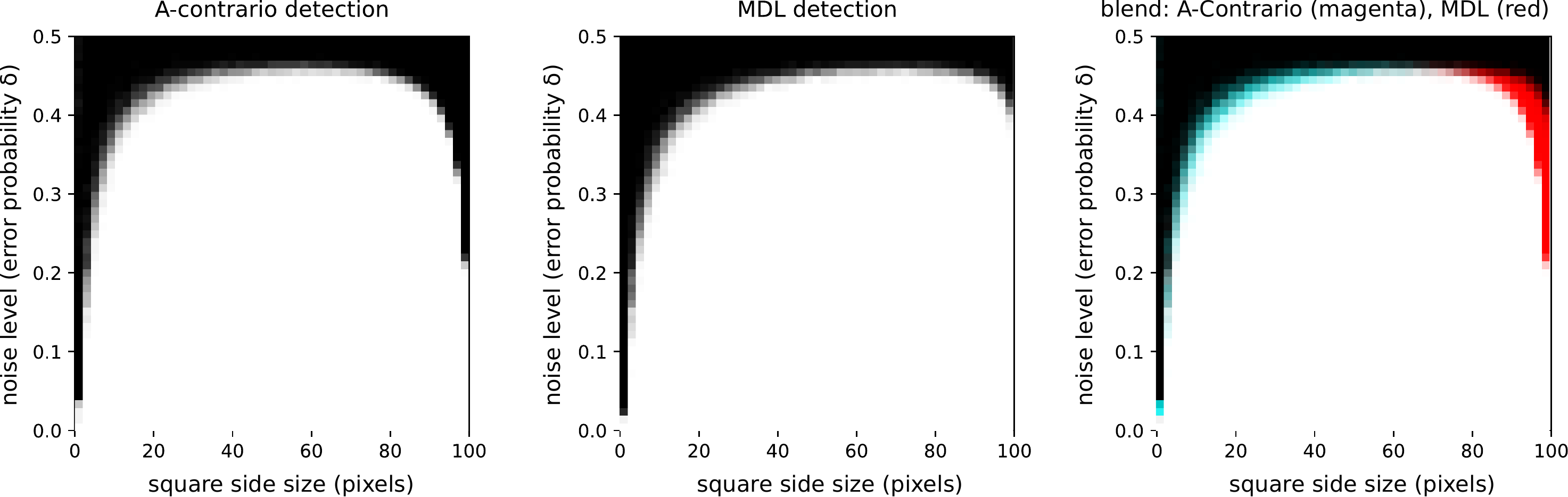}
\caption{Single square detection using MDL and a-contrario for different square
sizes (in terms of their side length) and noise level (expressed as probability
$\delta$ of a pixel being flipped from 1 to 0 and vice-versa).  For each noise
level and square size the experiment was repeated for 100 different noise
realizations.  In the first two images, the intensity of each pixel corresponds
to the percentage of cases where the corresponding method declared a detection
(white: always, black: never).  The last image is a color blend of the first
two: the red channel corresponds to the MDL image, and green and blue
channels (adding up to cyan) to a-contrario, so that  white regions (blue plus red plus green) 
correspond to places where both methods yielded a
detection).}
\label{fig:single-square-results}
\end{figure}

\Cref{fig:single-square-results} shows the result of a numerical experiment
illustrating how the scores of both criteria vary as a function of the size of
the square (horizontal axis) and the amount of noise (vertical axis).  The
results are quite similar up to the point where the area of the square is
roughly half of the image (size $\approx 70$ as the image is $100\times100$).
In this first part, as the size of the square grows, the density is better
evaluated and it is easier to distinguish $q_1$ from $q$; hence the detection
is dominated by the term $D(q_1||q)$ in both formulations.  The term
$D(q_0||q)$ in MDL is negligible at first as the background covers almost all
the image and $q_0\approx q$.  The a-contrario formulation is a bit more
sensitive than MDL; this is consistent with  the extra term in MDL
\cref{eq:mdl-devil-term-1}, which raises the detection threshold with respect
to that of the one obtained with a-contrario.

For sizes larger than 71, the square encompass more than half of all the
pixels.  Thus, $q$ is more and more affected by $q_1$, and thus less
distinguishable.  The term $D(q_1||q)$ gradually vanishes and the a-contrario
formulation is no longer able to detect the square.  For MDL, on the other
hand, as the square and the background have a symmetric role, the density $q_0$
becomes gradually distinguishable from $q$, and the term $D(q_0||q)$ becomes
dominant.  For the a-contrario approach, the square part is no longer anomalous
while MDL, analysing the full data, still sees a difference.

In summary, the numerical experiments show that both MDL and a-contrario result
in very similar detection criteria despite being strictly different from an
analytical perspective.

\section{Detection of multiple squares}\label{sec:squares}

In the previous section, we compared how the MDL and a-contrario approaches
behave in a simple scenario involving a detection task; the goal of this
section is to place both methods in a model selection problem.

In this case, instead of detecting the presence of individual squares, we will
be dealing with many squares (see~\cref{fig:multiple-squares-vs-error}).   All
possible combinations of squares are considered and the task is to decide the
number, size, and position of the squares that are deemed present.

Making the theoretically optimal decision according to both MDL and a-contrario
usually requires the evaluation of a very large number of events.  In practice,
this is not possible and both criteria need to be applied to a much smaller
subset of events using heuristics to propose relevant candidates.  The real-life applications presented later in this work fall into this category: as
the events involve the presence or absence of a large number of points or
segments in all possible positions, orientations, etc., some heuristic is
needed to narrow the choices. As the quality of these heuristics has an impact
on the results, care must be taken in comparing MDL and a-contrario approaches
under these circumstances. In this work, we sidestep this issue by using a
common heuristic to define the candidates to be tested by both methods.

In the present case, considering all the possible squares, with all their
possible sizes, even in a small image would result in a huge number of tests,
and the problem would become practically intractable.  For our purposes, it is
enough to compare a reduced set of arrangements. Concretely, we generate a
binary $256{\times}256$ image with a $2\times2$ array of small squares
separated by a given margin, which is a parameter.  Note that the four small
squares form a larger square whose size is always the same. As the margin
grows, the size of the smaller squares becomes smaller.  As before, the image
is contaminated by independent Bernoulli noise with error probability $\delta$.
Even in this case, there exist many possible explanations (each one of the
small squares or combinations of 2, 3, or all of them). Without loss of
generality, we 
reduce our candidate hypotheses to four: no detection, 
a single small square, four small squares, and the large square
formed by considering all
 four small squares as a single one. We expect the
choice to depend on the noise level and the proximity (margin) between the
smaller squares.

\subsection{MDL for multiple squares}\label{sec:squares:mdl}

The MDL detection scheme in this case is an extension of the procedure used for
a single square.  Each hypothesis consists of a number $c$ of squares, each one
described exactly as we did before for one square  (location, width, and
pattern of $0$s and $1$s within), plus a background formed by the pixels which
do not belong to any square, again described exactly as we did in the single
square scenario (\Cref{eq:square-l0}).

The number of pixels in the image is $n$. The $i$-th square has  $n_i$ pixels,
$k_i$ of which are $1$. The number of background pixels is  $n_0=n -
\sum_{i=1}^{c} n_i$, and the number of those which are $1$ is  $k_0 = k -
\sum_{i=1}^ck_i$. The corresponding empirical distributions are given by $q_i =
k_i/n_i, i=0,\ldots,c$.

The number of squares, $c$, is unknown beforehand, and so we must also describe
it to the decoder. We can do this in a number of ways. A simple one that will
usually do the work is an arbitrary geometric distribution, $P(c) \sim
(1/2)2^{-c}$, so that the additional term is just $-\log P(c) = c+1$. Summing all, the expression for $L_H$ is:
$$
   L_H = \log n_0 + \log {n_0 \choose k_0}
       + \sum_{i=1}^{c} \left[ (3/2)\log n + \log n_i
                               + \log {n_i \choose k_i} \right] + c + 1.
$$
As before, the hypothesis with the smallest $L_H$ is chosen.


\subsection{A-contrario for multiple squares}\label{sec:squares:nfa}

For the a-contrario formulation, we need to specify the family of tests, the
background model, and the statistic used to evaluate a test. We use the same
notation as before.

In this case, each test is determined by a set of squares defined inside the
image domain.  To take into consideration any number of squares, we can divide
the family of tests according to the number of squares and allocate an accepted
number of false alarms $\frac{\varepsilon}{2^c}$ to the sub-family of tests
containing $c$ squares.  We will set $\eta_i = 2^c N(c)$, where $N(c)$ is the
number of tests for $c$ squares.  Since $\sum_c \frac{1}{2^c}=1$, we know that
$\sum_i \frac{1}{\eta_i} = \sum_c\sum_{t\in N(c)}\frac{1}{2^c}
\frac{1}{N(c)}\leq1$; then, according to ~\cref{prop:nfa}, the quantity
$\eta_i\,\mathbb{P}(K \geq k)$ is an NFA.

Let us count the number of test in the sub-family of $c$ squares.  To simplify,
 we will neglect the 
squares overlapping and consider all
possible ways of selecting $c$ squares in the image.  Each square is determined
by its upper-left corner pixel 
and side length.  There are $n$ possible
choices for the upper-left corner and $\sqrt{n}$ choices for the square side
(this is  an upper bound, as some of the squares considered are not really
contained in the image domain).  There are about $n^{\frac{3}{2}}$ possible
squares in the image, and thus about $N(c)=\left(n^{\frac{3}{2}}\right)^c$
combinations of $c$ squares.  Again, this is an upper bound; but the important
thing is to get an estimation of the order of magnitude of the number of tests.
All in all, $\eta_i = 2^c \cdot n^{\frac{3}{2}c}$.

The background model $H_0$ considers that the pixels in the image are
independent and follow a Bernoulli distribution.  Its parameter $q$ is
estimated as the empirical density: $q=\frac{k}{n}$.

To evaluate a candidate $s$ with $c$ squares, we sum the total number of 1s in
all the squares $k(s)=\sum_{i\in s} k_i$ and the total number of pixels in all
squares, $n(s)=\sum_{i\in c} n_i$.  A candidate is considered a detection when
the total number of $1$s in $s$ is too large to what would be expected
according to the background model $H_0$.  Assuming that the squares composing
$s$ are not overlapping, then according to the background model $H_0$, the
number of $1$s would be $K(s)$, a random variable following the binomial
distribution of parameter $q$.  Then, the probability of observing $k(s)$ in
$H_0$ is:
$$
    \mathbb{P}( K(s) \geq k(s) ) = B\Big(n(s), k(s), q\Big)
$$
where $B(n,k,q)$ is the tail of the binomial distribution:
$$
  B(n,k,q) = \sum_{i=k}^n {n \choose k}q^i(1-q)^{n-i}.
$$
Finally, as $k(s) = \sum_i k_i$ and $n(s) = \sum_i n_i$, the NFA is:
$$
  \mathnfa = 2^c \cdot n^{\frac{3}{2}c}
             \cdot B\left(\sum_i n_i,\, \sum_i k_i,\, \frac{k}{n}\right).
$$
As usual, a detection is declared when $\mathnfa\leq\varepsilon$, with
$\varepsilon=1$. As mentioned in \cref{sec:notation}, the $\mathnfa$ score can
also be used to select the best configuration: the configuration with the
smallest NFA is the least expected one under $H_0$ and is therefore the
preferred one.

\subsection{Comparison on multiple squares}\label{sec:squares-comparison}

In this case, the comparison will be limited to numerical experiments.

\begin{figure}[p]
\centering
\begin{subfigure}{\textwidth}
\centering
\includegraphics[width=.8\linewidth]{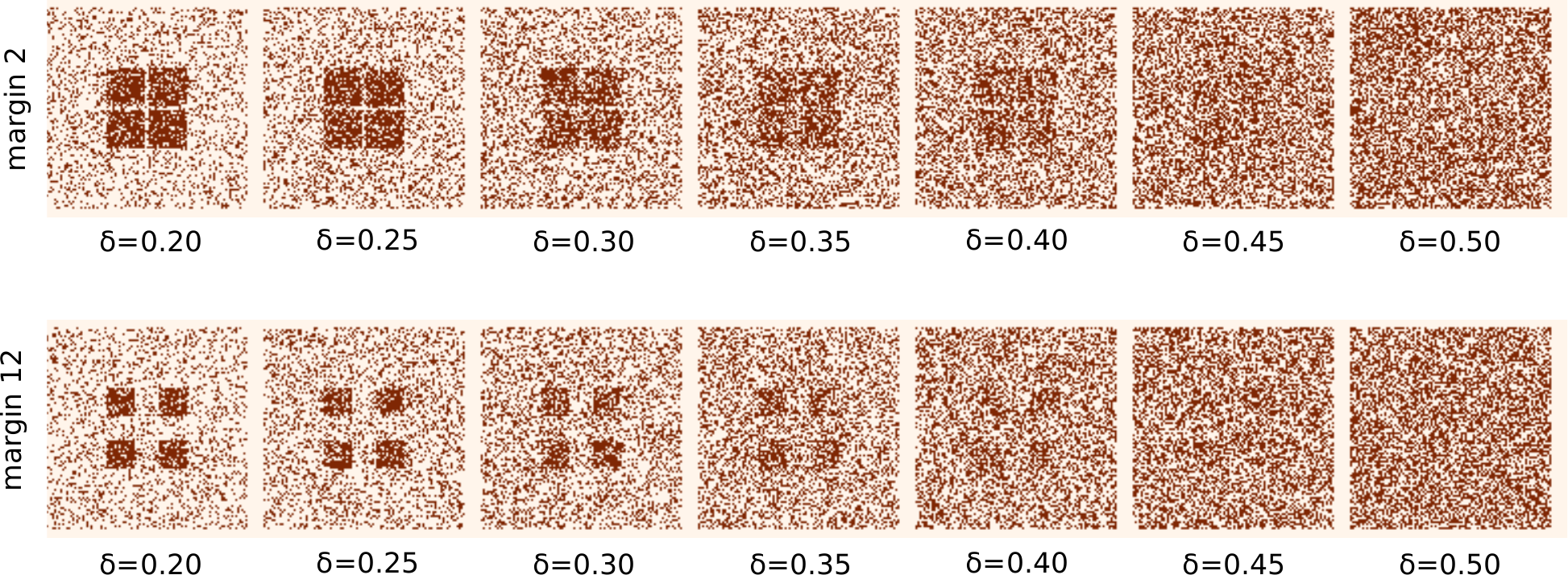}\\
\end{subfigure}
\begin{subfigure}{.4\textwidth}
\includegraphics[width=\linewidth]{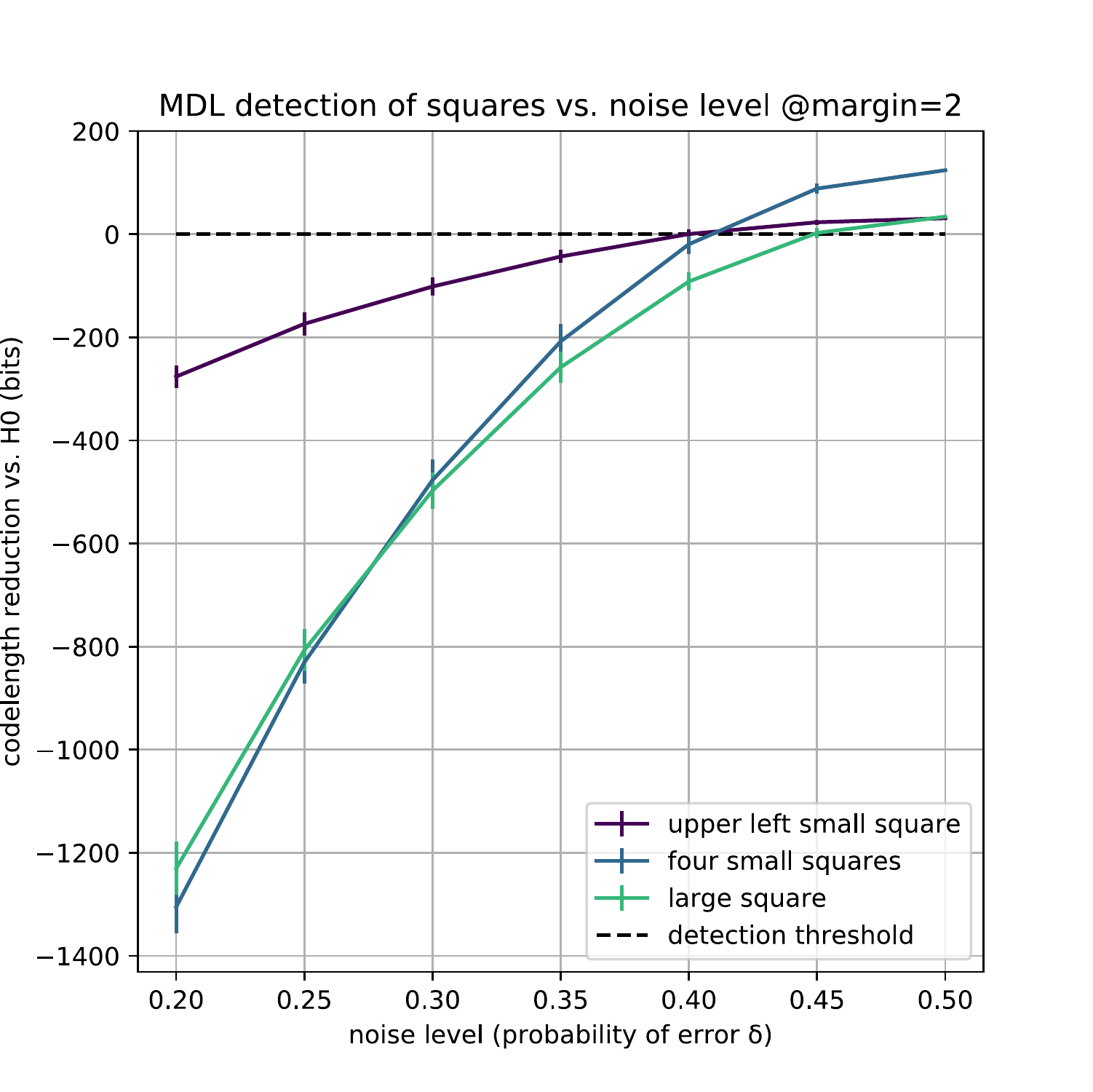}
\end{subfigure}
\begin{subfigure}{.4\textwidth}
\includegraphics[width=\linewidth]{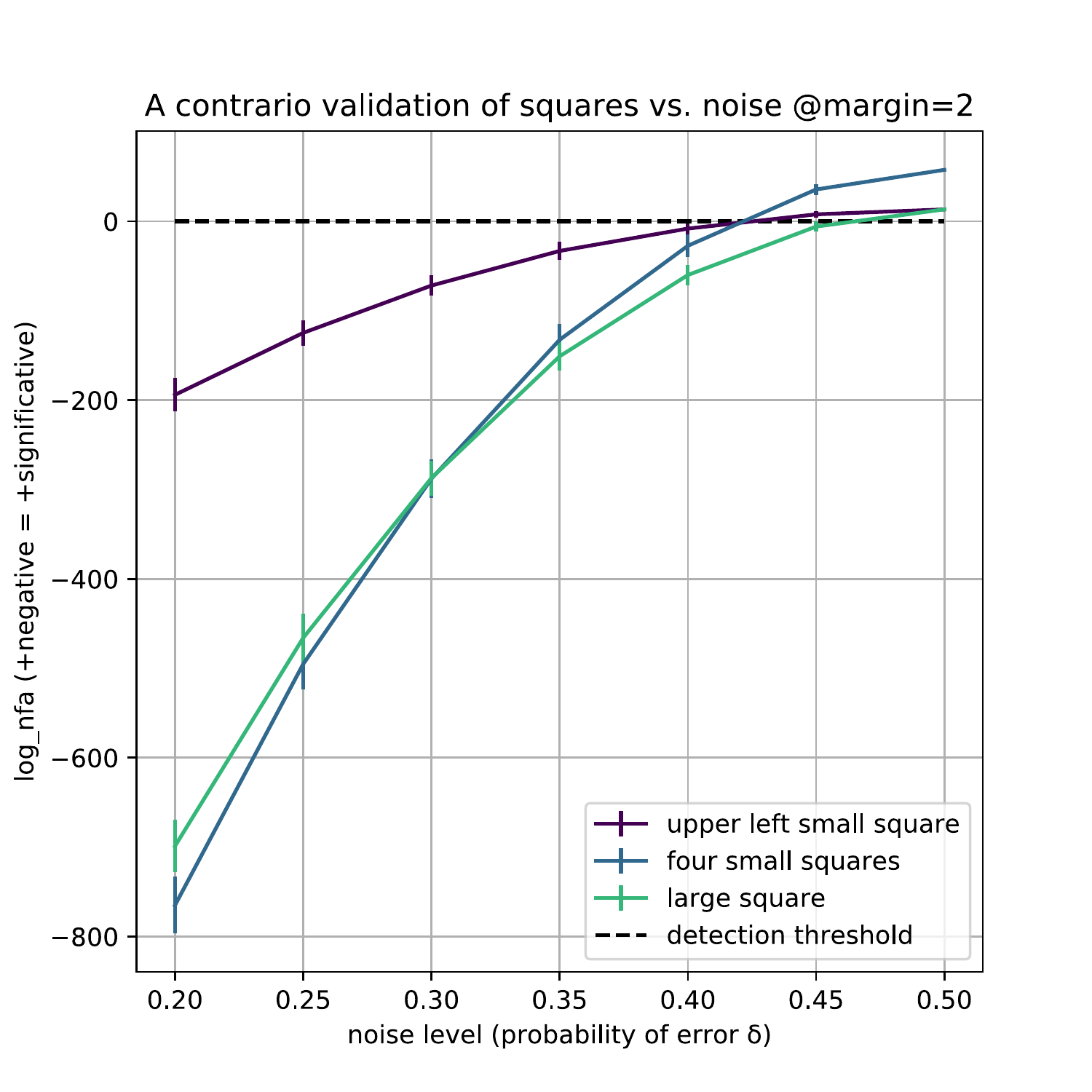}
\end{subfigure}
\begin{subfigure}{.4\textwidth}
\includegraphics[width=\linewidth]{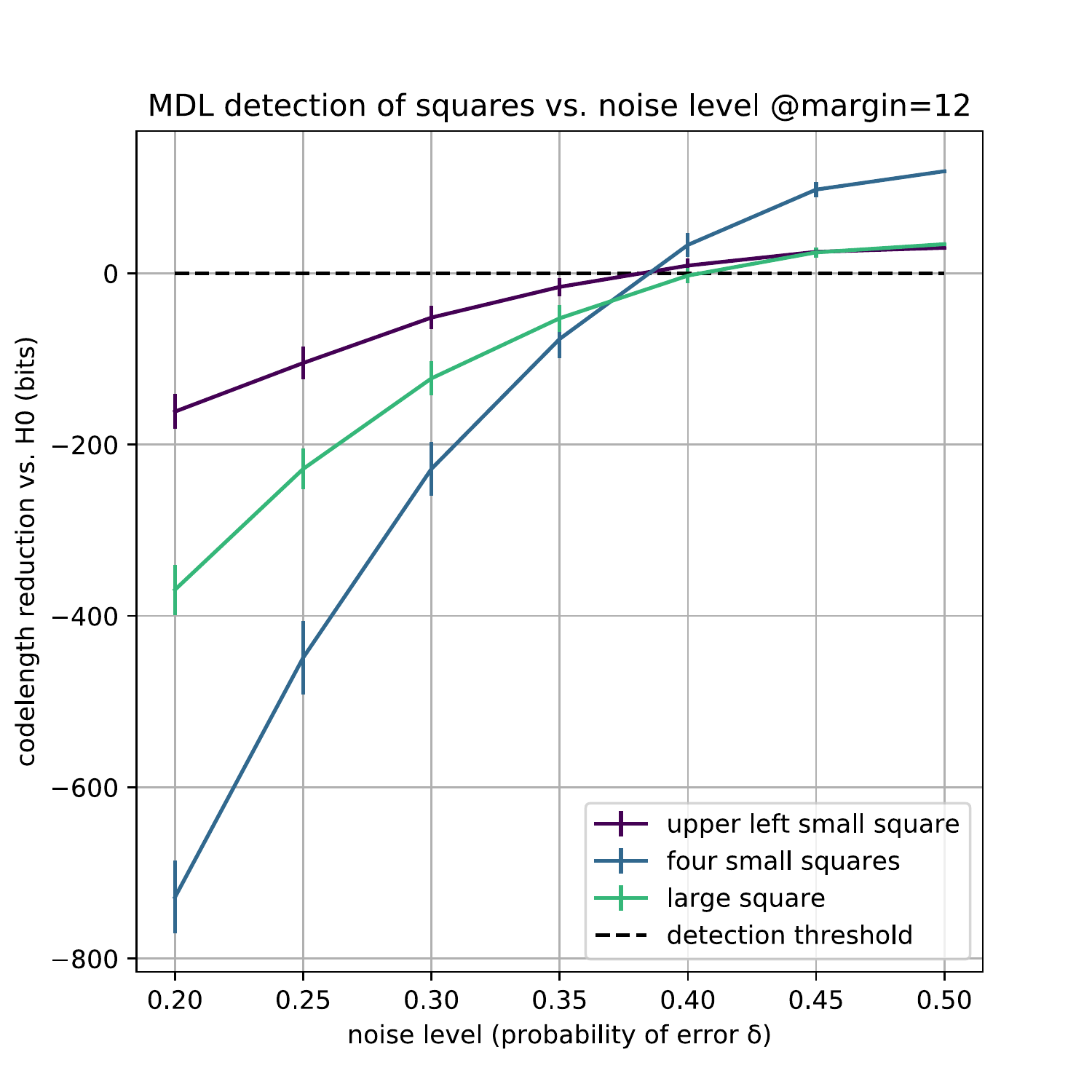}
\end{subfigure}
\begin{subfigure}{.4\textwidth}
\includegraphics[width=\linewidth]{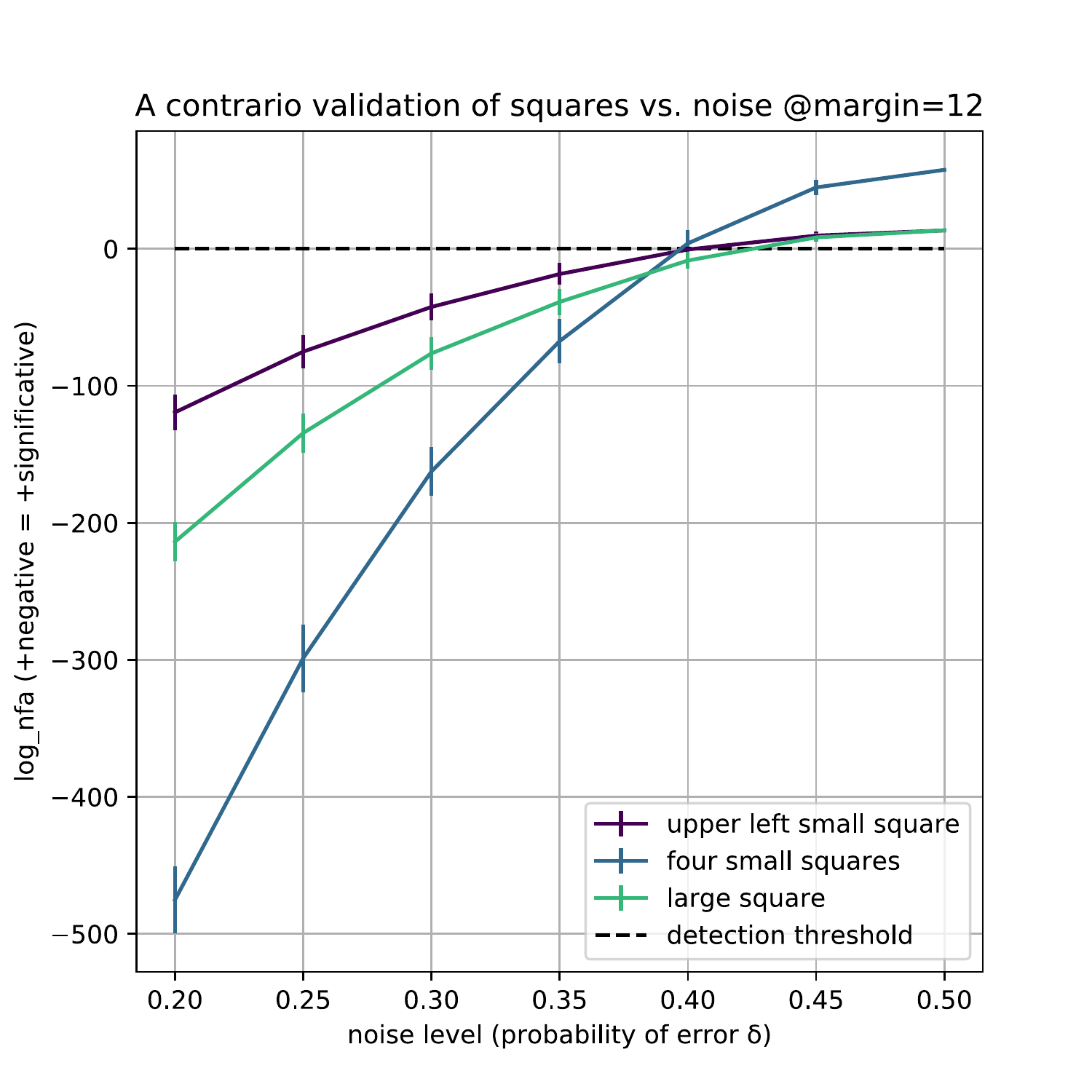}
\end{subfigure}
\caption{Detection of multiple squares as a function of the error probability,
for fixed margins. First and second rows: sample images for this setting.
Third row:  MDL and a-contrario results for the small margin case; a detection
occurs when the score is below $0$. Fourth row: results for the large margin
scenario. In the a-contrario method, lower scores indicate more significant
events. See the text for a discussion of these results.}
\label{fig:multiple-squares-vs-error}
\end{figure}

\paragraph{Sensitivity as a function of noise level}
\cref{fig:multiple-squares-vs-error} shows the detection results of both MDL
and a-contrario on the multiple squares case, for two different margin values
(small and large), as the noise level increases.

\begin{figure}[p]
\centering
\includegraphics[width=0.90\textwidth]{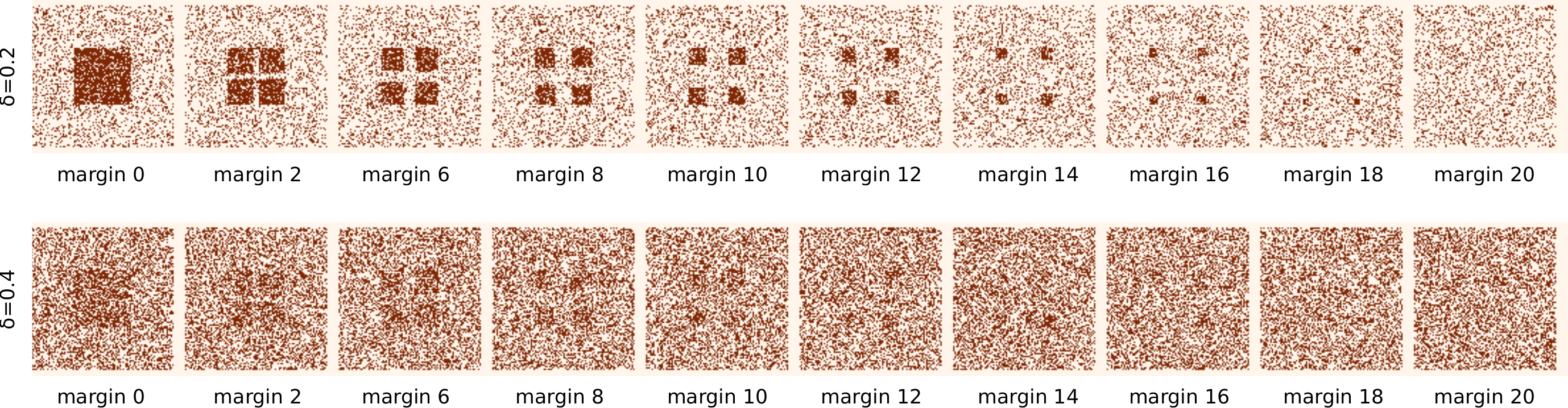}\\
\includegraphics[width=0.45\textwidth]{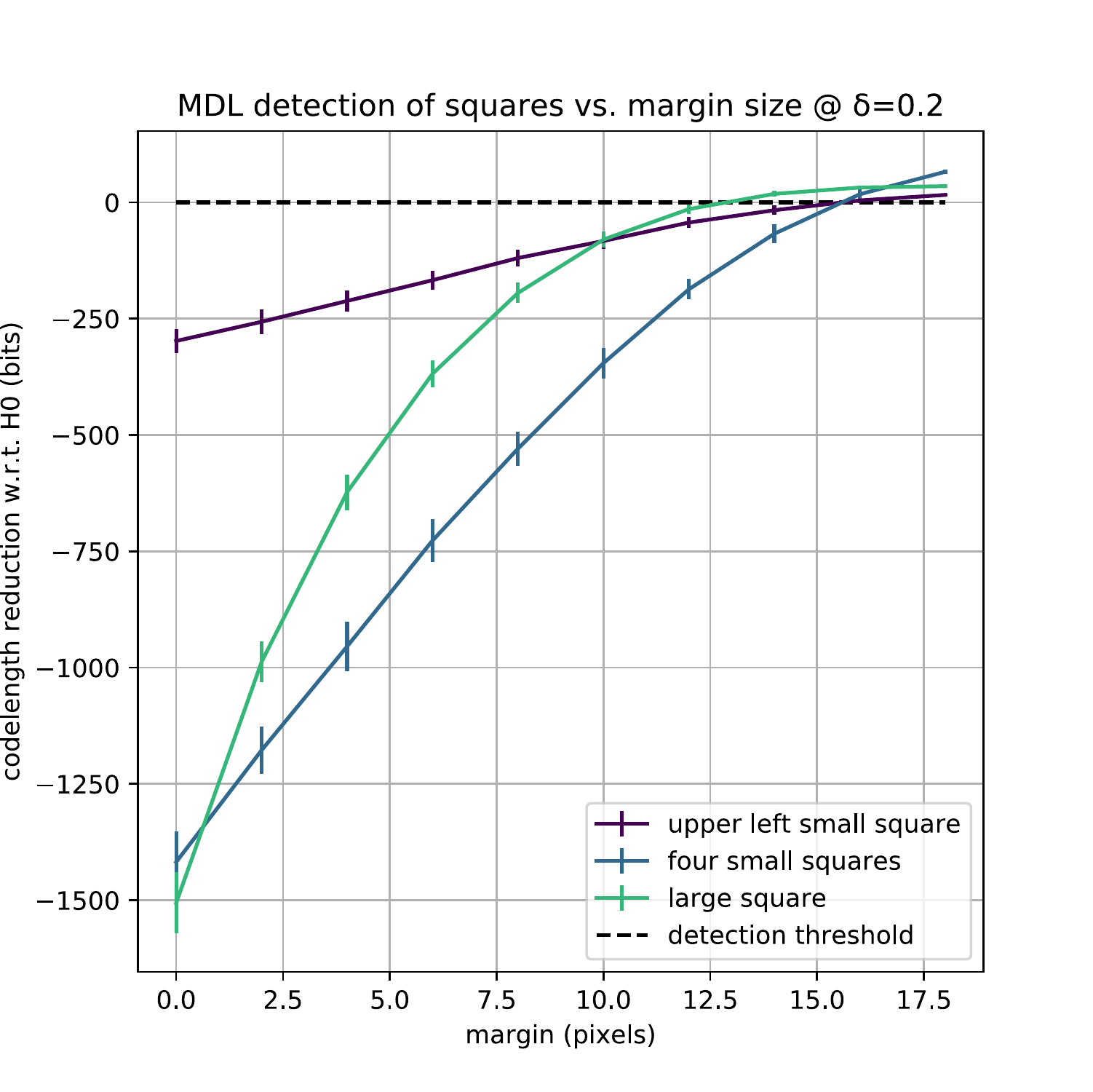}
\includegraphics[width=0.45\textwidth]{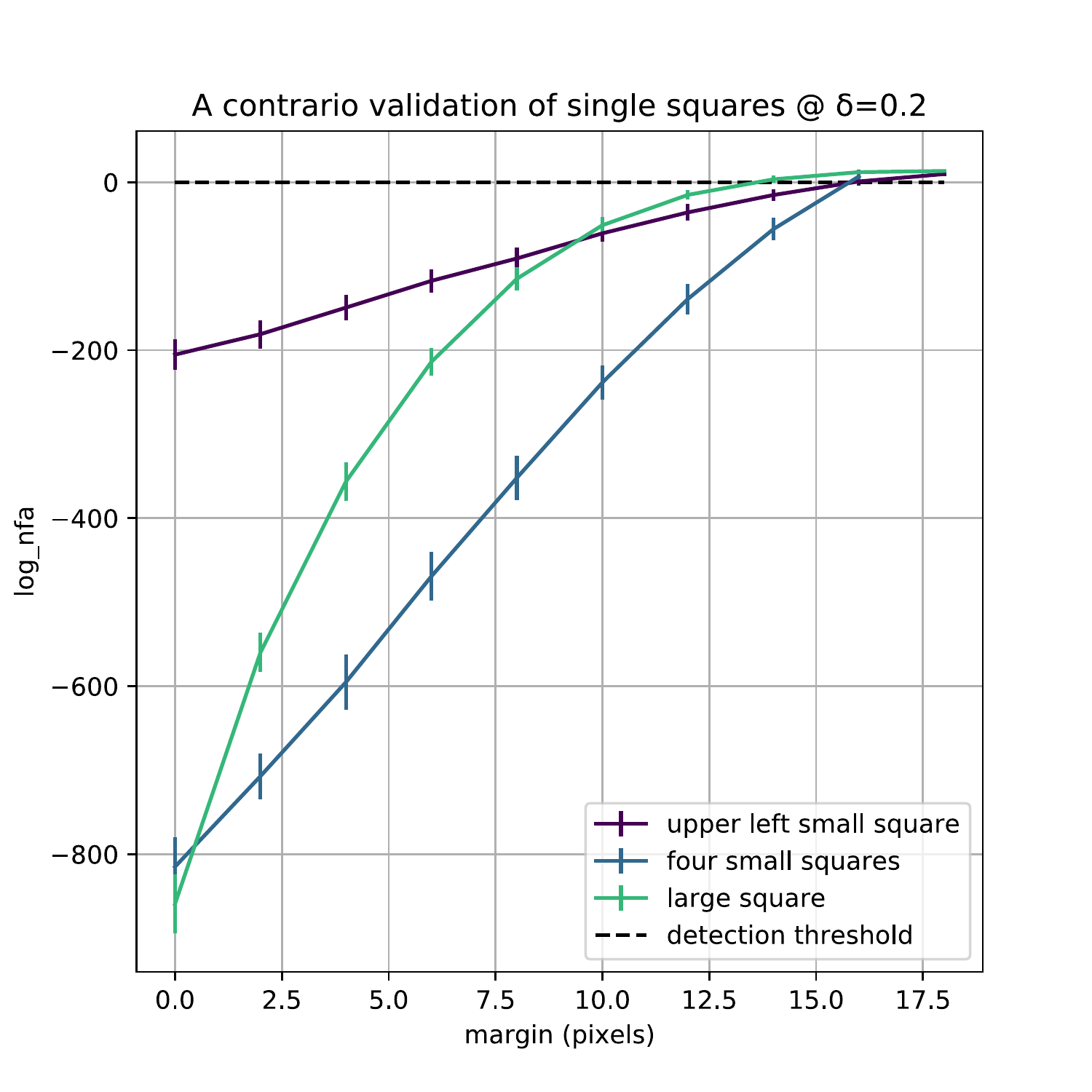}\\
\includegraphics[width=0.45\textwidth]{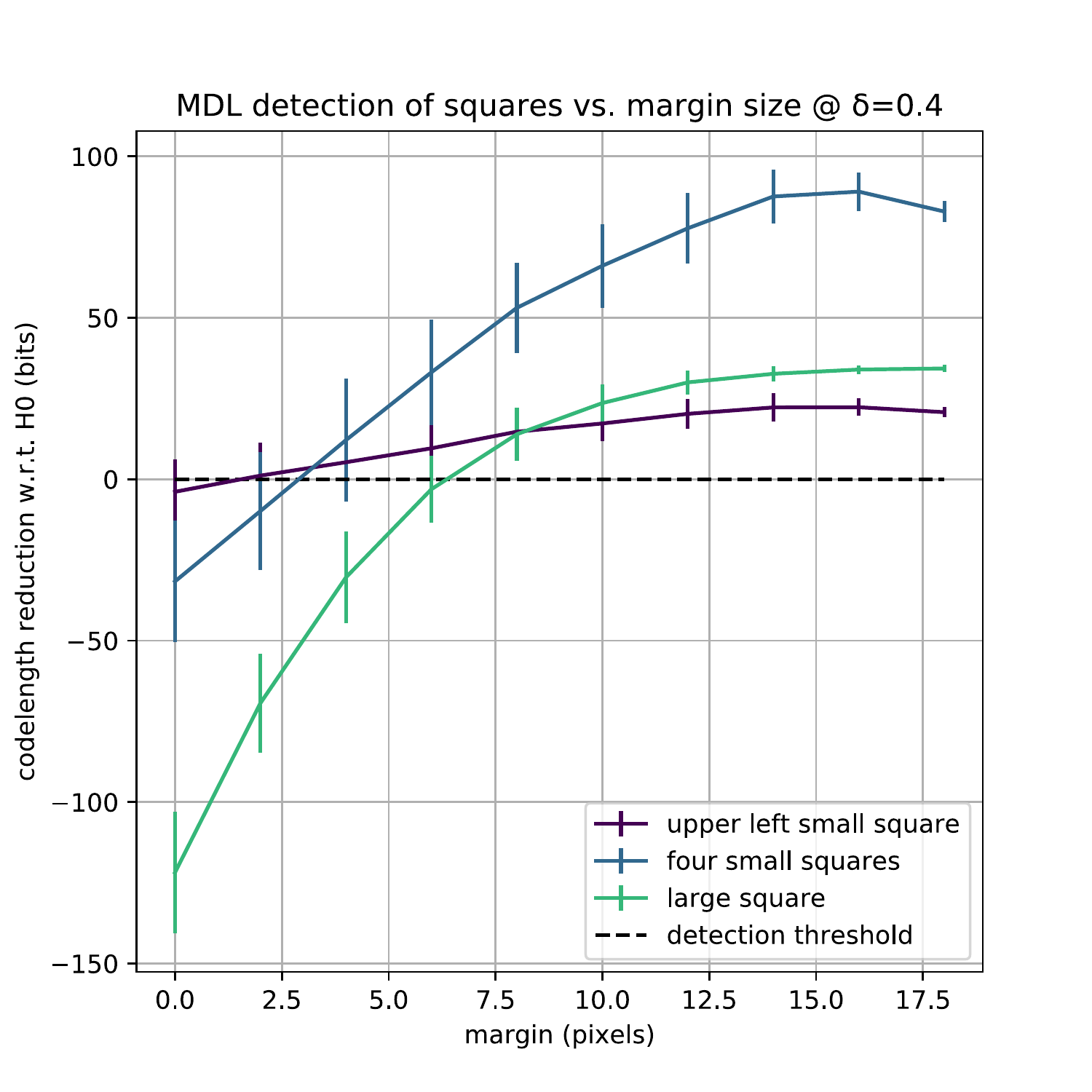}
\includegraphics[width=0.45\textwidth]{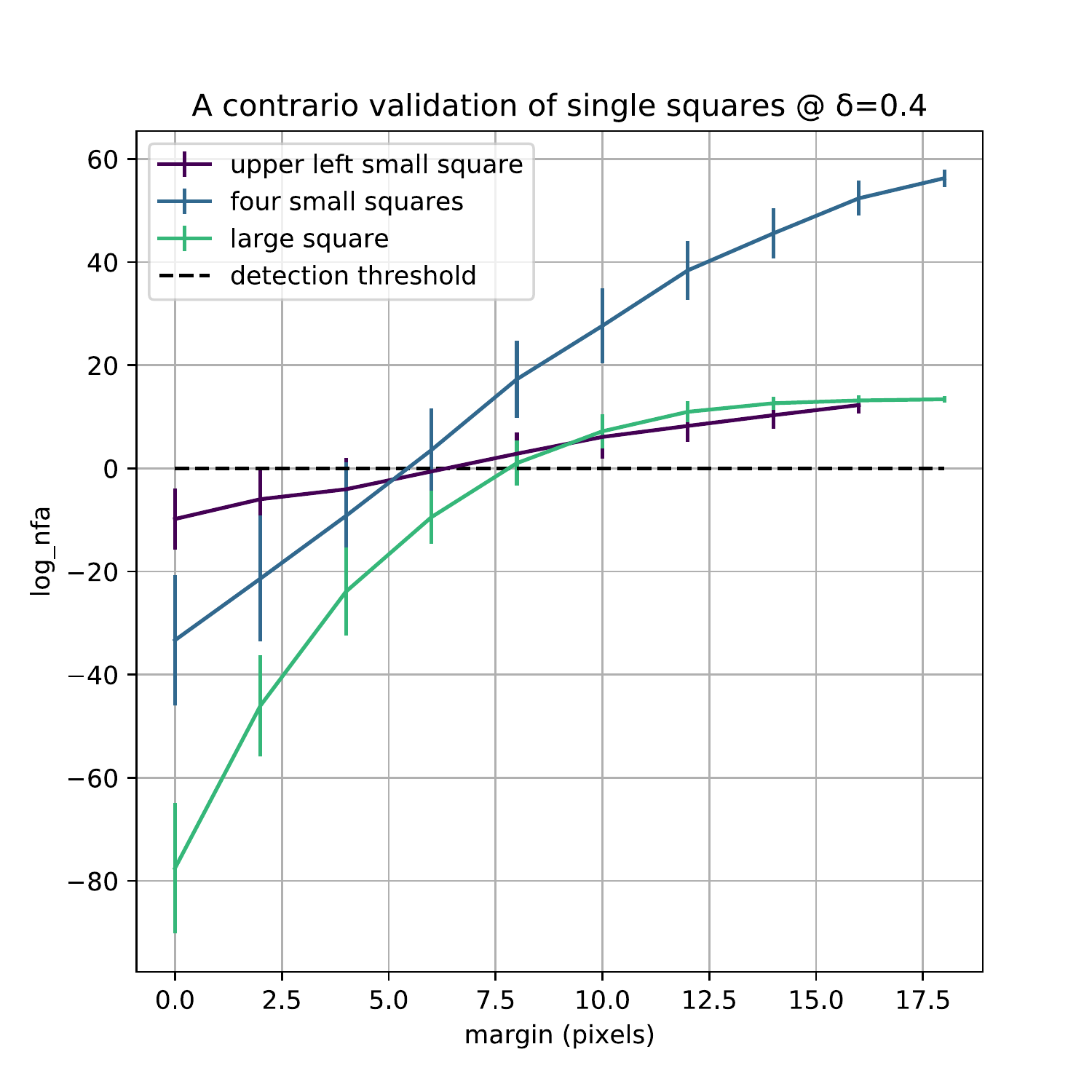}
\caption{Detection of multiple squares as a function of the margin, for a fixed
error rate $\delta$. The top row shows images of four squares separated by
different margins  under mild ($\delta=0.2$) noise.  The second row shows
similar images under high ($\delta=0.4$) noise. Third row shows the MDL (left)
and a-contrario (right) results for $\delta=0.2$. A value below $0$ indicates a
positive detection.  Last row shows the experiment output for $\delta=0.4$. See
text for a discussion.}
\label{fig:multiple-squares-vs-margin}
\end{figure}

For large margins, both methods produce the correct result (four small squares)
up to a noise level of about $\delta=0.4$ (MDL actually stops detecting a
little earlier, at about $\delta=0.38$). These thresholds correspond to the 5th
image from the left of the second row of \cref{fig:multiple-squares-vs-error};
after a visual inspection of those images, it can be argued that the automatic
threshold of $\delta=0.4$ is in agreement with human perception.\footnote{Of
course, a thorough assessment of such a statement would require a carefully
crafted visual perception study, which is out of the scope of this paper.}

For a smaller margin (2 pixels), again, both methods behave almost identically.
Interestingly, both methods switch to the simpler model (large square) for
noise levels $\delta \geq 0.3$; in this case, it is less clear which would be
the reasonable limit with a casual inspection of the corresponding image
(\cref{fig:multiple-squares-vs-error}, first row, 3th image from the left).

\paragraph{Sensitivity as a function of margin size}
\cref{fig:multiple-squares-vs-margin} shows the results of both methods on the
multiple squares experiment, this time while varying the size of the margin
between the small squares.  The experiment is repeated for low ($\delta=0.2$)
and high ($\delta=0.4$) noise levels.

For the low noise setting, both methods give very similar results, with
identical detection thresholds at a margin of size $16$; this corresponds to
the next-to-last image on the first row of
\cref{fig:multiple-squares-vs-margin}, which again seems to indicate that the
results of both methods agree with the human perceptual threshold for this
case.

In the (very) high noise setting, as can be seen in
\cref{fig:multiple-squares-vs-margin}, it is clear that \emph{both methods
switch to the simpler explanation} of a single, large square, in all cases.
Here a-contrario is again slightly more sensitive, detecting a square for a
margin of up to $8$ pixels, whereas MDL stops doing so at a margin of $6$
pixels. The a-contrario threshold seems to be more in line with human visual
perception in this case (\cref{fig:multiple-squares-vs-margin}). This could
offer some sort of validation to the a-contrario method, which is inspired (to
some extent) by the human visual system. We discuss this idea in a slightly
more formal way in \cref{sec:discussion}.

\section{Approximation of shapes using polygons}\label{sec:poly}

In this experiment, we are presented with a closed curve that separates a
foreground object from the background in a binary image.  The curve is
approximated by several connected dots defining a polygon, and the task is to
simplify the polygonal approximation by removing unnecessary points.  Ideally,
we would like to select the optimal subset of vertices from all the possible
subsets of initial points, but this quickly becomes computationally infeasible.
This is a particular instance of the well-known, general problem of
\emph{subset selection} in statistics.  As in the general case, one must resort
to a manageable subset of candidate point subsets.  Typically, two alternatives
exist to construct those subsets: \emph{backward stepwise selection} (BSS) and
\emph{forward stepwise selection} (FSS)  (see \cite[Chapter~3]{hastie08}).

\newcommand{\iter}[1]{^{#1}}

In BSS, the starting subset $P\iter{0}$ is the full set of candidate points
$P$.  An initial fitness score $\phi\iter{0}$ is associated to $P\iter{0}$.
Let $|A|$ denote the size or cardinality of the set $A$.  At step $t$,
$|P\iter{t}|$ candidate subsets are produced by removing each one of the
$|P\iter{t}|$ points in $P\iter{t}$, and their corresponding scores are
computed.  If the score of the best candidate is smaller than $\phi\iter{t}$,
then that candidate is declared to be the new best solution, $P\iter{t+1}$, and
its score is assigned to $\phi\iter{t+1}$.  The process stops at step $t$ if no
candidate yields a score better than $\phi\iter{t}$, or $|P\iter{t}|=3$ (the
smallest number of points in a non-degenerate polygon).

The FSS heuristic goes in the opposite direction: the general algorithm starts
with an empty subset and greedily adds the best candidate in each iteration
until the score $\phi\iter{t}$ can no longer be improved. In our case, the FSS
algorithm would require us to begin with at least \emph{three} points from $P$
(this can already be a problem if $P$ is large), and add points from $P$  until
the addition of new points to $P\iter{t}$ does not improve the current score
$\phi\iter{t}$.

Below we develop both a-contrario and MDL-guided BSS heuristics for choosing
the best subset of points from the full polygon, which is obtained using the
Devernay Sub-Pixel Edge Detector~\cite{devernay-ipol}. As the name suggests, this is a model selection problem, for which MDL is
naturally suited.  As before, we describe both approaches in detail and then
proceed to compare their results on a number of selected cases.

\subsection{MDL approach for polygon approximation}\label{sec:poly:mdl}

As before, we are presented with a 2D binary image $\vx$ of $n$ pixels.  The
polygon is defined by an ordered set $P$ of $c$ vertex coordinates $(i,j)$. The
interior of the polygon (which includes the vertices themselves) is denoted by
$I$; the outside is defined by a set $O$.  This problem is largely analogous to
that of the single square problem presented in \cref{sec:square}, the only
difference being the description of the region itself, which is a polygon
instead of a square. Thus, we follow the same general encoding strategy: for a
given candidate polygon $P\iter{t}$ the code-length associated to it is
$L(\vx,P\iter{t})=L(\vx|P\iter{t})+L(P\iter{t})$. The first term is the
concatenation of two enumerative codes: one for $I\iter{t}$, and one for
$O\iter{t}$. Again, we define $n_1$ to be the number of pixels of the polygon
in $I\iter{t}$, $k_1$ the number of 1s inside the polygon in $I\iter{t}$,
$k_0=k-k_1$, and $n_0=n-n_1$. Note that $n_0$, $k_0$, $n_1$ and  $k_1$ change
for each iteration $I\iter{t}$ (to simplify the notation, the super-index $t$
was not added to $n_0$, $k_0$, etc).  We get:
$$
  L(\vx|P\iter{t}) = \log n_1 + \log {n_1 \choose k_1}
                   + \log n_0 + \log {n_0 \choose k_0}.
$$

The interesting part is  how to describe $P\iter{t}$, as there are different
possibilities depending on various assumptions we may make about the polygon
which could allow us to encode the coordinates in a clever way (e.g.,
differentially).  As before, for the sake of simplicity, we will describe the
2D coordinates of $P\iter{t}$ as if they were independent. This requires $\log
n$ bits per coordinate, so that $L(P\iter{t})=c\log n$.  Finally, we also need
to describe the number of vertices, $c$, which we do as we did in
\cref{sec:squares:mdl}, requiring $c+1$ additional bits.  The overall
code-length is very similar to \cref{eq:mdl-square-l1}:
\begin{equation}\label{eq:mdl-polygon}
  L(\vx,P\iter{t}) = 1 + c(1+\log n) + \log n_1 + \log {n_1 \choose k_1}
                   + \log n_0 + \log {n_0 \choose k_0 }.
\end{equation}
Notice that the term $c+1$ was combined with $c\log n$ as $1+c(1+\log n)$,
which also emphasizes that the cost of describing the number of vertices,
$c+1$, is negligible for typical image sizes $n$ (e.g., $\log (256{\times}
256)=16 \gg 1$).  The trade-off in \cref{eq:mdl-polygon} is simple: removing a
vertex from $P\iter{t}$ will save us $1+\log n$ bits.  On the other hand, this
will introduce errors in the frontier between $I\iter{t}$ and $O\iter{t}$ so
that the empirical distributions of $I\iter{t}$ and $O\iter{t}$ will deviate
further from their true underlying Bernoulli distributions. A well-known result
from Information Theory establishes that this will result in a longer overall
code-length with overwhelming probability~\cite[Chapter~2]{cover06}.

\subsection{A-contrario approach for polygon approximation}

As with MDL, the a-contrario formulation for this case is mostly analogous to
the square detection problem: given the foreground shape, the evaluation of its
significance relies on the empirical density of its interior pixels.  The main
difference is that the family of tests includes all polygonal curves in the
image domain instead of all squares.

As in the case of multiple squares, the set of tests is decomposed into
subsets, each corresponding to a given number of sides in the polygon; the
accepted number of false alarms for the subset of tests which consider polygons
with $s$ sides is set to $2^s$. Now, given $s$, a crude approximation to the
number of possible polygons with $s$ sides is to assume that any image pixel can be a vertex, which gives us $n^s$ possible polygons. Combining both factors, we obtain:
$$
   \eta_s = 2^s \cdot n^s.
$$
As before, $n_1$ is the number of pixels inside the polygon, $k$ the total
number of 1s in the image,  $k_1$ the number of 1s inside the polygon, and the
empirical density is $q=\frac{k}{n}$. Under the null hypothesis, a Bernoulli
process with $P(x=1)=q$, the probability of observing at least $k_1$ 1s among
the total $n_1$ pixels inside the candidate polygon is given by
$$
   \mathbb{P}(K \geq k_1) = B\left(n_1,k_1,q\right)
$$
and the NFA is given by
$$
   \mathnfa = 2^s \cdot n^s \cdot B\left(n_1,\,k_1,\,q\right).
$$
When $\mathnfa<\varepsilon$ the event is considered $\varepsilon$-meaningful
and the candidate is validated.

The NFA is by construction a quantity to decide the presence or not of a
pattern.  However, it can also be used to decide between alternative
interpretations of the same data.  When a given structure is indeed present in
the data, a candidate similar to the actual structure, for example sharing most
of the pixels, may also result in a meaningful test.  Nevertheless, the actual
structure would probably be the one with the largest deviation from the
background model.  This deviation from the background model is measured by the
NFA.  Thus, the candidate with the smallest NFA often corresponds to the actual
structure.  As a consequence, selecting the candidate with smallest NFA can be
used as a model selection procedure.

\begin{figure}[htbp]
\centering
\includegraphics[width=0.32\textwidth]{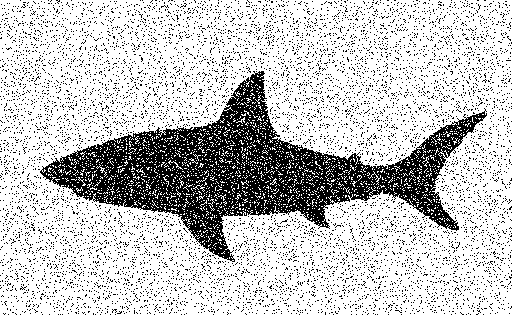}
\includegraphics[width=0.32\textwidth,height=1.65in]{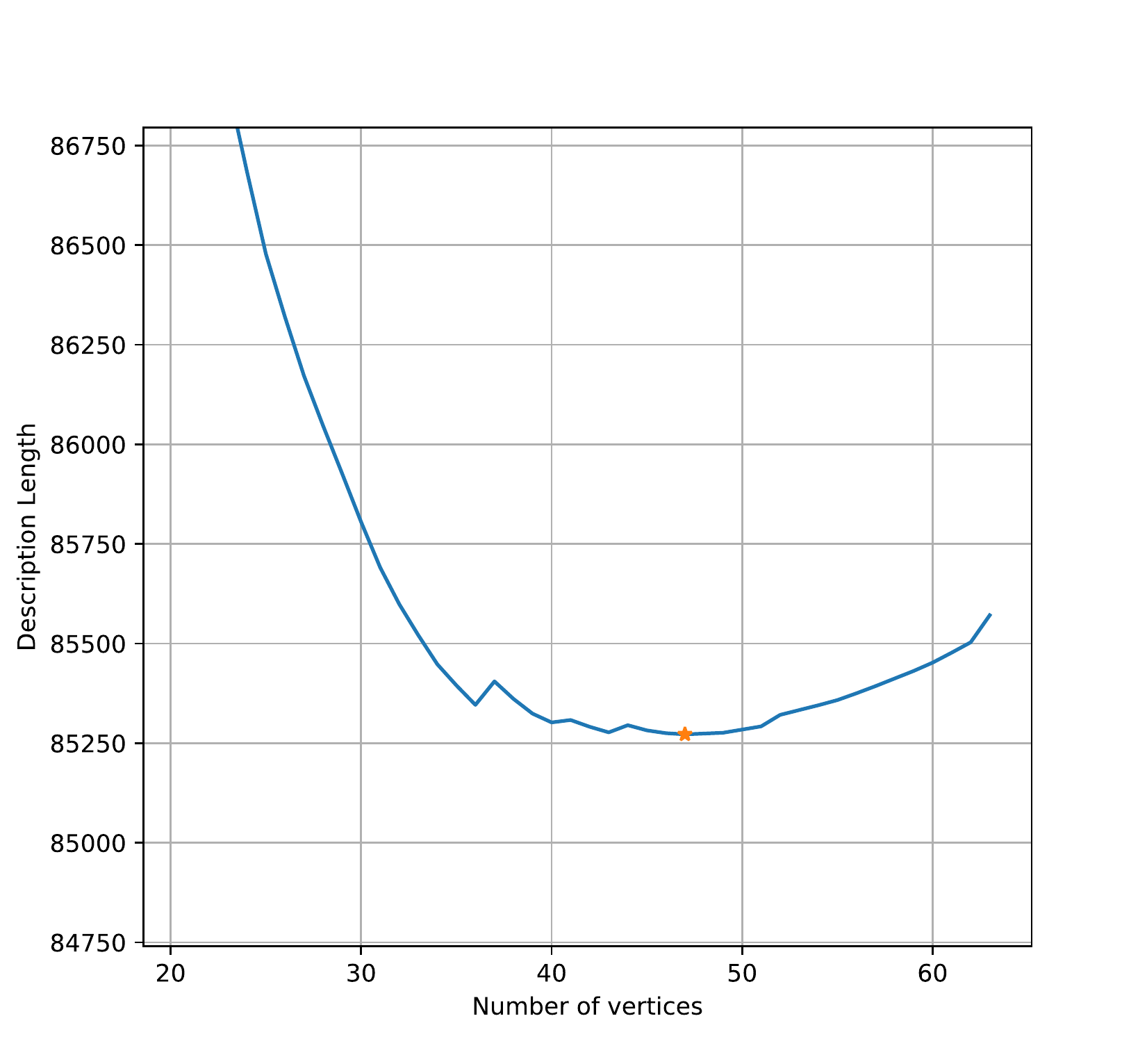}
\includegraphics[width=0.32\textwidth,height=1.65in]{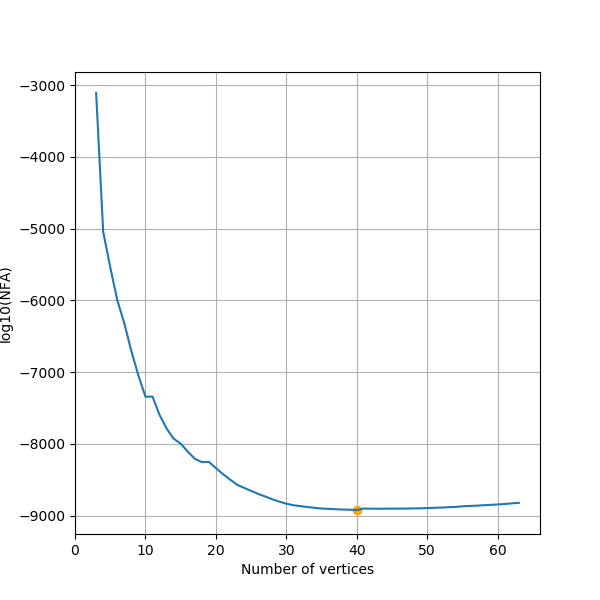}\\
\includegraphics[width=0.32\textwidth]{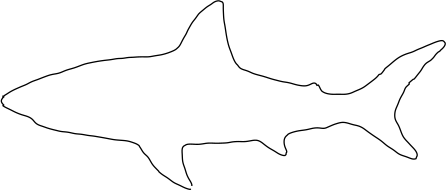}
\includegraphics[width=0.32\textwidth]{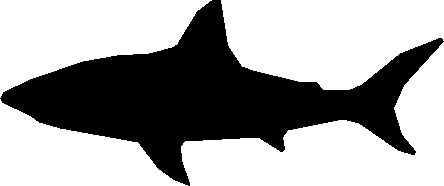}
\includegraphics[width=0.32\textwidth]{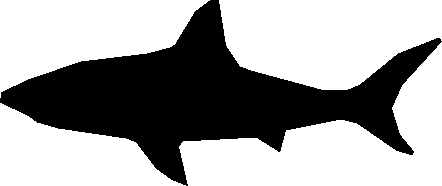}\\
\caption{Polygon approximation comparison.  Top left: noisy image;  bottom
left: full polygon estimated using the Devernay
algorithm~\cite{Devernay95anon-maxima} (63 vertices); top center: MDL
code-length vs. polygon size (the best is marked with a star); top right:
$\log_{10}(\mathnfa)$ vs. number of polygon vertices (the best is marked with a
star); bottom center: best MDL polygon ($47$ vertices); bottom right: best
a-contrario  polygon ($40$ vertices).}
\label{fig:poly-comparison}
\end{figure}

\begin{figure}[htbp]
\centering\includegraphics[width=0.99\textwidth]{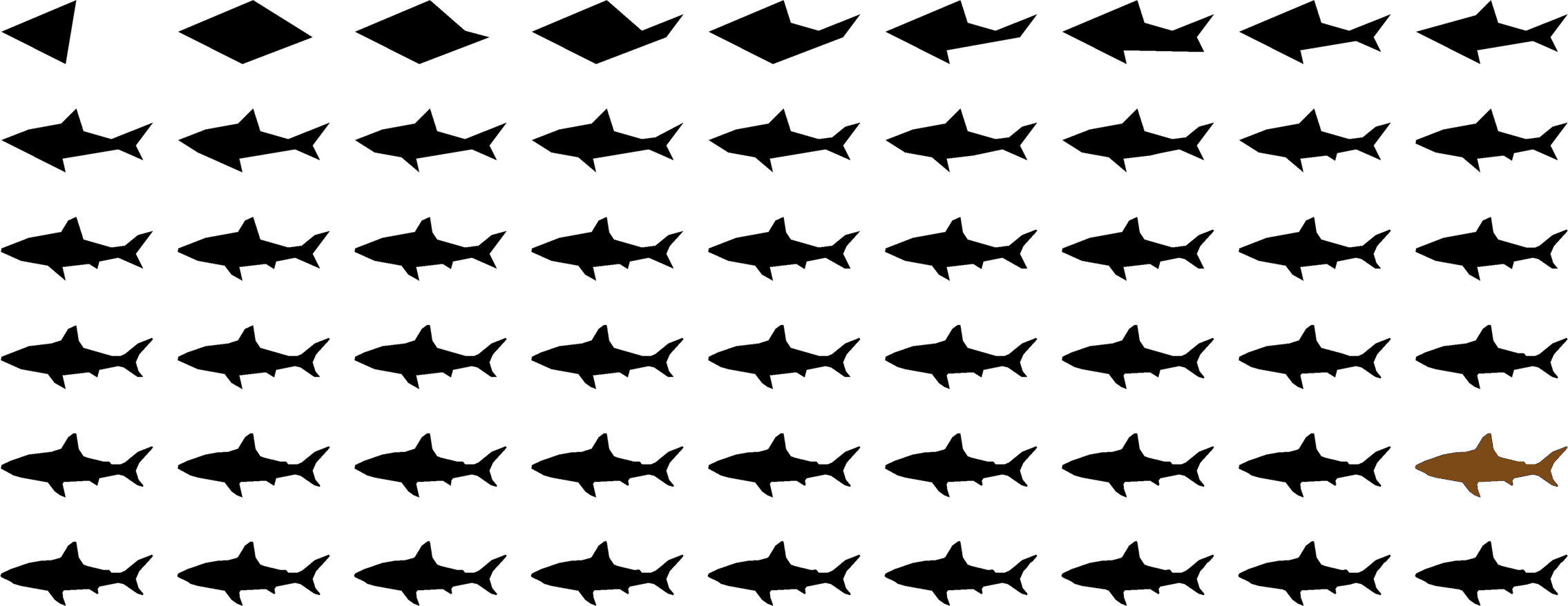}
\caption{Sequence of candidate polygons obtained using BSS 
Selection 
guided by MDL. The polygon sizes range from $3$ to $56$. The chosen
one is shown in gold ($47$ vertices).}
\label{fig:poly-mdl}
\end{figure}

\begin{figure}[htbp]
\centering\includegraphics[width=0.99\textwidth]{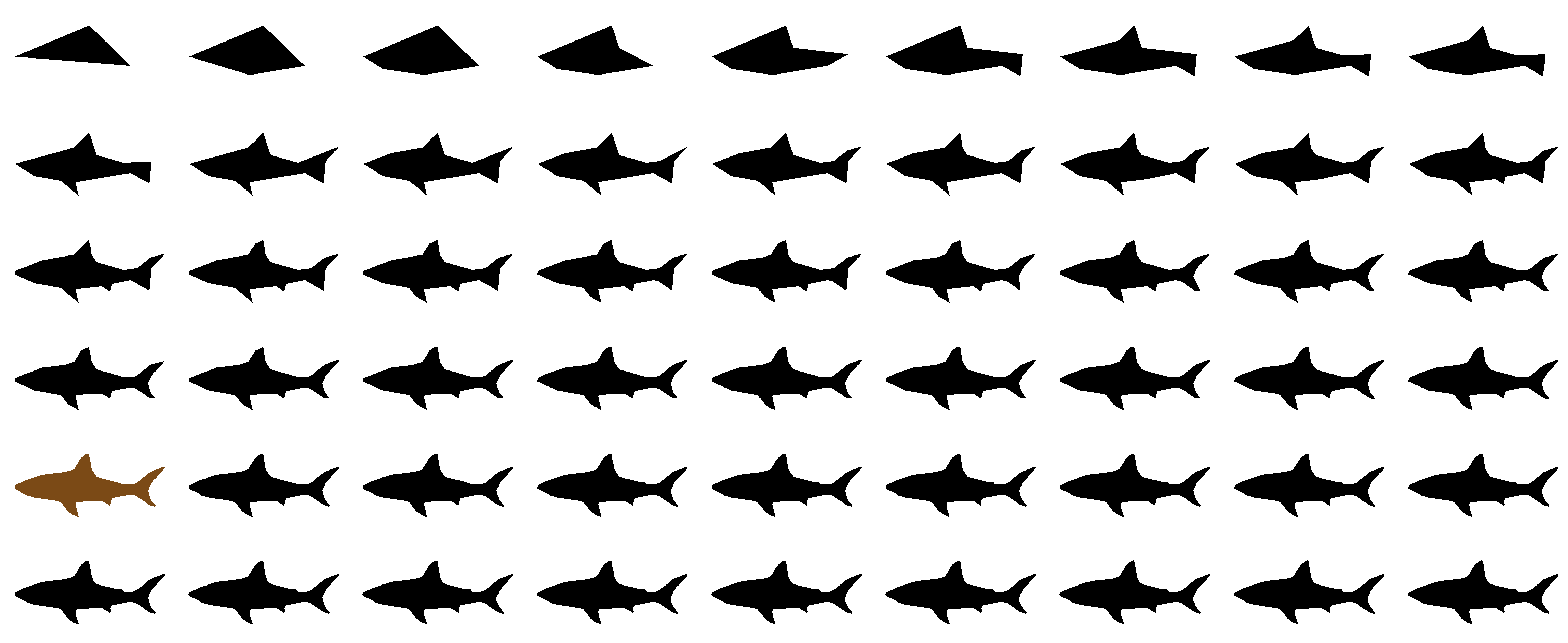}
\caption{Sequence of candidate polygons obtained using BSS 
Selection 
guided by NFA value in the a-contrario approach. The polygon sizes
range from $3$ to $66$. The chosen one  is shown in gold ($40$ vertices).}
\label{fig:poly-nfa}
\end{figure}

\subsection{Comparison on polygon approximation}

\Cref{fig:poly-comparison} illustrates the results obtained. Despite their
differences, both methods produce similar results: the ``score-vs-number of
points'' curves are similar in shape, the minima are close, and the
approximated shapes are similar too.  \Cref{fig:poly-mdl} and
\cref{fig:poly-nfa} show the sequence of candidates obtained using the BSS
heuristic guided by MDL and a-contrario respectively; comparing these figures,
the differences between the BSS paths obtained using each method can be
observed.  Our aim is not to find which one has a ``better'' result (something
difficult to define), but to show that both approaches can be used for a real
model selection problem and that the solutions are similar.

\section{Line segments in 2D images}\label{sec:lsd}

The aim of this final task is to detect the solid line segments present in an
image.  The LSD algorithm~\cite{lsd,lsd-ipol} is a popular and successful
application of the a-contrario approach for this problem.  Below we describe
the original a-contrario-based LSD algorithm,  develop its MDL counterpart, and
compare the results of both algorithms on a number of examples.

LSD algorithm is based on the orientation of the image gradient.  A fast
heuristic is applied to group neighbor pixels which share a similar gradient
orientation, producing a segmentation of the image into candidate regions.  A
rectangle is fit to each region, resulting in candidate line segments; these
candidates are then evaluated independently using an NFA metric: those
meaningful enough are kept, and the rest  discarded.  In order to apply MDL to
this case, we need to formulate such detection sub-problem as a data
compression one.  Below we describe both approaches in detail, beginning with
the original a-contrario method.

\subsection{A-contrario line segment detection}

\def\vd{\mathbf{d}}

Here we summarize the main ideas behind the LSD algorithm. We refer the reader
to the original work~\cite{lsd} for further information, and to~\cite{lsd-ipol}
for implementation details, in particular those pertaining to the heuristic
search for candidates.

Given an initial image $\vx$ with a total of $n$ pixels, the input to this
method is an image gradient orientation map $\vd \in [-\pi,\pi)^n$.  The null
hypothesis $H_0$ for this problem  assumes that the gradient orientations are
independent and isotropic, that is, $d_{ij} \sim
\mathrm{Uni}([-\pi,\pi))\;\forall  i,j.$

The family of tests is composed of all candidate rectangles in the image; a
rectangle $r$ is the subset of image coordinates determined by the triad
$(a,b,w)$, where $a$ and $b$ are the endpoints of the line segment that splits
the rectangle in half along its shorter dimension (that is, the ``center line''
of the rectangle), and $w$ is the \emph{width} of the rectangle, its shorter
dimension. We also define $\lambda_r$ to be the normal direction of the segment
$(a,b)$.

As with the other examples in this work, a heuristic is applied to produce a
reduced set of feasible candidates to be tested for meaningfulness;  the
validation step is agnostic of this heuristic, assuming that all  possible
rectangles in the image, up to pixel precision, are tested.

For an image with $n$ pixels, the number of possible pairs of endpoints is
bounded above by $n^2$, and the width of the rectangles by $\sqrt{n}$, yielding
an upper bound for the number of tests of $N = n^{5/2}$. Despite its crudeness,
using this approximation has proven good enough for practical purposes.

What remains is to define a criterion for deciding whether a particular
candidate is declared as a positive detection, or discarded. There are many
possible ways of doing so: below we describe the method used
in~\cite{lsd,lsd-ipol}.

Given a candidate $r$, a positive detection is declared if the number of pixels
in $r$ aligned with $r$ is too high to be produced by chance.  A pixel $x_{ij}
\in r$ is considered aligned if $|d_{ij} - \lambda_r| \leq \rho$ for some
pre-defined tolerance $\rho \geq 0$.

Let $n_r=|r|$ be the number of pixels in $r$, and  $0 \leq k_r \leq n_r$ be the
number of such pixels that are aligned.  Following the a-contrario formalism,
we define the NFA for this problem as
\begin{equation}
  \mathnfa(r) = N \cdot \mathbb{P}\Big[ K_r \geq k_r \Big],
\end{equation}
and declare a detection if $\mathnfa(r) \leq \varepsilon$. Here the random
variable $K_r$ represents the number of aligned pixels in a rectangle $r$ for
the null hypothesis $H_0$.

Under $H_0$, the gradient orientation map is assumed isotropic and the
probability of a single pixel being aligned with $r$ is $\mathbb{P}(|d_{ij} -
\lambda_r| \leq \rho) = \theta =\frac{\rho}{\pi}$.  Furthermore, as the angles
in $\vd$ are assumed independent, the aforementioned events are themselves
independent and $K_r$ is the sum of $n_r$ independent Bernoulli random
variables.  Thus, $K_r$ follows a Binomial distribution of parameter $\theta$,
\begin{equation}\label{eq:p-lsd}
   \mathbb{P}\Big[ K_r \geq k_r \Big] = B\Big(n_r, k_r, \theta\Big)
\end{equation}
where $B(n_r,k_r,\theta)$ is the tail of the binomial distribution. Note that,
as expected,
$$
  \mathbb{P}\Big[ K_r \geq k_r \Big] \rightarrow 0
$$
as $k_r$ increases, which corresponds to increasingly more meaningful events.

To reduce the impact of the particular value used for the tolerance $\rho$, the
LSD algorithm uses multiple values\footnote{See~\cite{lsd-book} for a slightly
different a-contrario formulation not requiring a tolerance $\rho$.}. Formally,
each different value of $\rho$ defines a new test.  If $\gamma$ is the number
of different $\rho$ values  used to evaluate each candidate, the grand total of
tests must be corrected by this factor: $N =  n^{5/2} \cdot \gamma.$

In summary, the NFA of a rectangle $r$ with $n_r$ pixels, $k_r$ of which are
aligned, is given by:
\begin{equation}\label{eq:lsd-nfa}
   \mathnfa(r) = n^{5/2} \cdot \gamma \cdot B\Big(n_r, k_r, \theta \Big).
\end{equation}
As usual, we fix $\varepsilon=1$; all  rectangles $r$ for which  $\mathnfa(r) <
1$ are considered detections.

\subsection{MDL line segment detection}\label{sec:mdl-lsd}

The object on which the line segment detection is performed in LSD is not the
original image $\vx$ itself, but its gradient orientation map $\vd$.  In order
to apply the MDL criterion to this problem, it is this data that we shall
encode.  Furthermore, in LSD, the NFA criterion is applied independently to a
set of candidate rectangles based only on the values of $\vd$ and the normal of
the rectangle $r$, $\lambda_r$.  Accordingly, in order to produce a meaningful
comparison between both criteria, we shall apply MDL to the problem of
compressing $\vd$ within each candidate rectangle $r$.

The modular design of the public LSD implementation~\cite{lsd-ipol} allows us
to quickly  test the proposed MDL variant by simply replacing the NFA criterion
with the MDL one. This also provides a common set of candidates to work with,
so that the results are easier to analyze and compare solely in terms of the
criteria themselves.

The encoding problem for this case is as follows. We need to describe the
gradient orientation map $\vd$.  According to the LSD pipeline, each element of
$\vd$ can be computed as a function of three pixels from $\vx$, each taking on
$2^8$ possible values\footnote{The formula in~\cite{lsd-ipol} is a function of
$4$ values, but it can be reduced to $3$.}.  Correspondingly, each $\vd_{ij}$
can take on $2^{24}$ possible values.

We are already given a set of candidate \emph{rectangles} which may contain a
significant portion of points aligned with their respective normal directions
$\lambda$.  As with LSD, we declare a point $(i,j)$ to be aligned with its
corresponding rectangle if $|d_{ij}-\lambda_r| \leq \rho$, where $\rho$ is a
threshold to be defined.  The points that do not belong to any rectangle are
ignored and are assumed to be described using an uniform distribution on the
$2^{24}$ possible angle values.

Let $n_r$ be the number of points in a rectangle $r$ and $\vd_r$ be the subset
of elements of $\vd$ indexed by the coordinates in $r$.  In the MDL paradigm, a
rectangle $r$ will be declared detected if we obtain a shorter description
length by assuming them to belong to the rectangle $r$, rather than to the
background. Formally, let $L(\vd_r,r)$ be the description length obtained if we
assume that the points in $r$ form a line segment, and $L(\vd_r,H_0)$ be the
one obtained if they are assumed to be background pixels.  Note that, under the
uniform assumption, the latter is simply $L(\vd_r,H_0)=24n_r.$

For $L(\vd_r,r)$, as before, we have to consider two pieces of information: the
description of the rectangle itself, $L(r)$, and the description of the aligned
points given $r$, $L(\vd_r|r)$. The simplest description of $r$, although
somewhat wasteful, is to describe its two endpoints and its width. Each
endpoint requires $\log n$ bits. As in LSD, we bound the width by $\sqrt{n}$.
This gives us a total of $\frac{5}{2}\log n$ bits per rectangle, which is 
the number of tests $N$ in the a-contrario framework (before the
correction factor $\gamma$ applied due to the different angle thresholds).

A rectangle contains aligned points, unaligned points, and points whose
gradient is considered undefined by the gradient computation algorithm.  Our
strategy is to encode the aligned points with one distribution, and the
unaligned and undefined points with another. In order to describe the subset of
aligned points we resort once again to an Enumerative Code as the one used
in~\cref{sec:square} and~\cref{sec:squares}.  If $n_r$ is the total number of
points in the rectangle and $k_r$ is the number of those that are aligned, we
require $\log n_r + \log {n_r  \choose k_r}$ bits to indicate which of them are so.
The unaligned/undefined points are described as background samples using $24$
bits each, for a total of $[n_r-k_r]24$ bits.

As for the aligned points, we know that $|d_{ij}-\lambda_r| \leq \rho$.  A
simple and conservative hypothesis in this case is that the angles $d_{ij} \sim
\mathrm{Uni}[\lambda_r-\rho,\,\lambda_r+\rho]$.  Now, as the $2^{24}$ possible
angles are approximately  uniformly distributed on $[-\pi,\pi]$, we expect
about $(\rho/\pi)2^{24}$ possible values to fall within
$[\lambda_r-\rho,\,\lambda_r+\rho]$.

In summary, for a rectangle $r$ with $n_r$ points, $k_r$ of which are aligned,
we obtain:
$$
   L(\vd_r,r) = \frac{5}{2}\log n + \log n_r + \log {n_r \choose k_r}
              + 24(n_r-k_r) + (24+\log \rho/\pi)k_r.
$$
Noting that $24(n_r-k_r) + (24+\log \rho/\pi)k_r = 24n_r + k_r\log \rho/\pi$,
the  MDL score is given by:
\begin{equation}\label{eq:lsd-mdl}
    L(\vd_r,r)-L(\vd_r,H_0) = \frac{5}{2}\log n + \log n_r
                            + \log {n_r \choose k_r} + k_r\log (\rho/\pi).
\end{equation}
As $\rho < \pi$, the last term, which represents the reduction in code-length
due to the tighter distribution of the aligned samples, is strictly negative.
The other terms, all related to the description of the
rectangle, are all non-negative. Thus,  intuitively, the decision rule of
\cref{eq:lsd-mdl} will deem a rectangle significant if enough points $k_r$ are
aligned so that the savings outweight the cost of describing the rectangle.
Note also that the term $\log {n_r \choose k_r}$ is positive but diminishes
with $k_r$ (and becomes $0$ when $k_r=n_r$, that is, all points are aligned), 
reinforcing the evidence of alignment as $k_r$ grows.

As a final comment, notice that we have not added a term for the number of
segments in the image. This is because, contrary to the previous examples, we
are considering the detection of each segment as an independent test, instead
of considering the set of segments in the image as a whole.  In terms to be
explained in \Cref{sec:equivalence}, it work by parts.

\begin{figure}[htp!]
\centering
\includegraphics[width=0.49\textwidth]{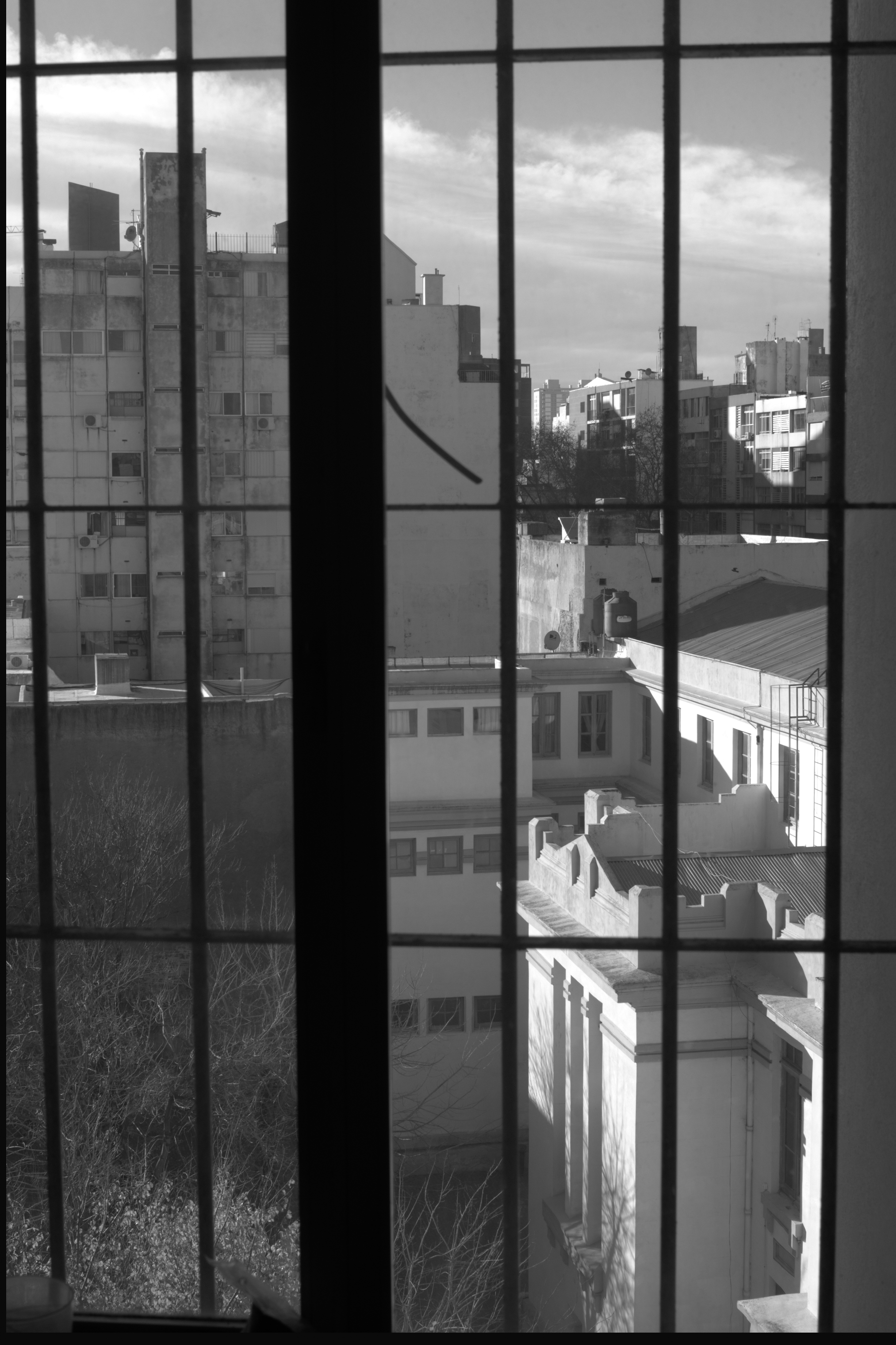}
\includegraphics[width=0.49\textwidth]{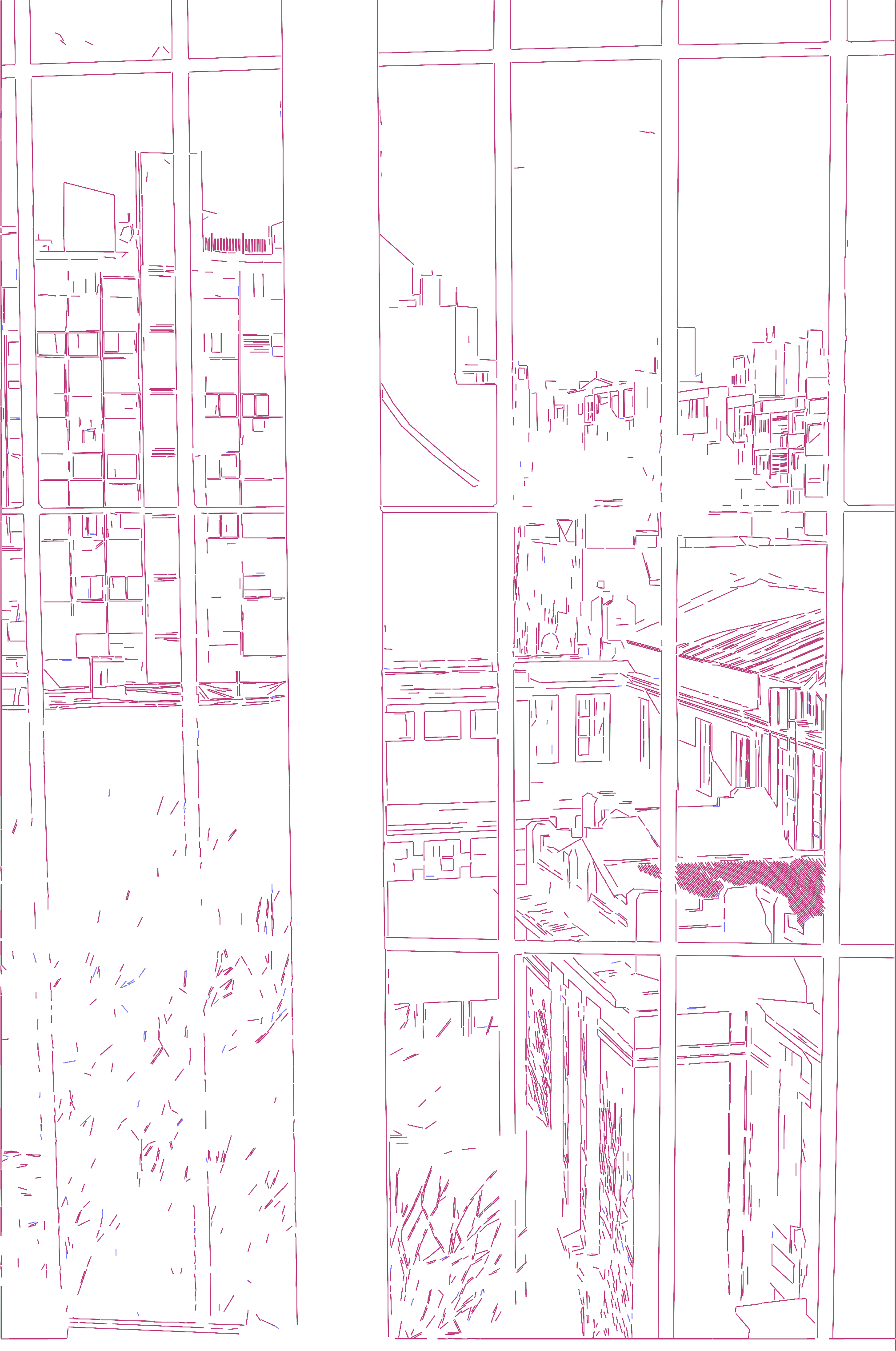}
\caption{Visual comparison of line segments detected using a-contrario (blue)
and MDL (red) approaches; segments detected by both are shown in violet.  As
can be observed, both methods yield extremely similar results (this is a high
resolution image; zoom in to see details).}
\label{fig:cuadra-comp}
\end{figure}

\begin{figure}[htp!]
\centering
\includegraphics[width=\textwidth]{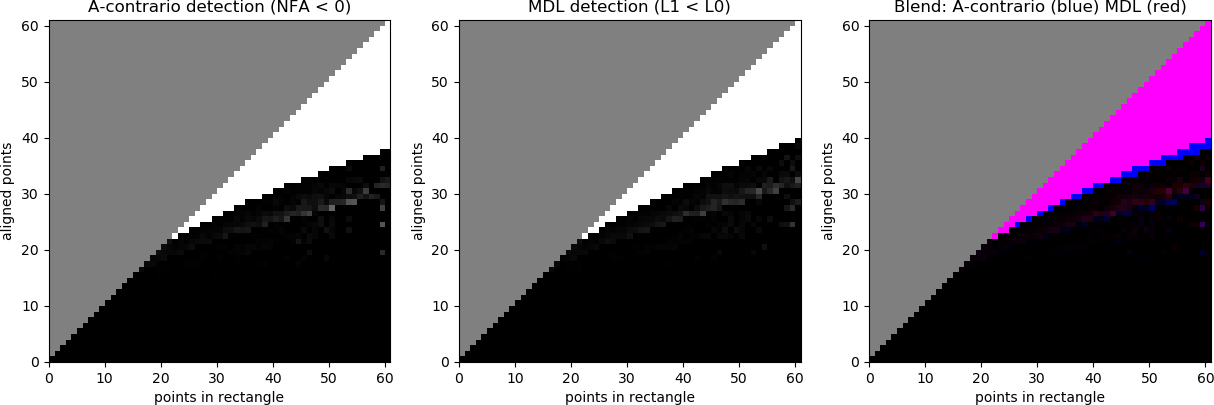}
\caption{Line Segment Detection probability as a function of the number of
points within a candidate rectangle, and the number of those points which are
aligned. The size of the rectangle is cut at $60$ (the largest value
for which we have samples along the whole vertical axis), and of course the
number of aligned points cannot exceed the size of the rectangle (this region
is filled with gray in all three images). Again both criteria produce very
similar results.}
\label{fig:cuadra-comp-2}
\end{figure}

\subsection{Comparison on line segment detection}

Here, as in the polygon approximation case (\Cref{sec:poly}), our aim is to
show that it is possible to use both approaches for a real detection problem
and that the solutions are similar.  \Cref{fig:cuadra-comp} shows a visual
comparison of both detection criteria on a sample image. Line segments detected
using a-contrario are marked in blue and the ones detected by MDL in red.
Segments which were detected by both are shown in violet.  As can be seen, both
methods yield extremely similar results.

In order to analyze the detection capability of both methods we plot the
probability of detection for different combinations of rectangle size and
number of aligned points on a large number of test images.
\cref{fig:cuadra-comp-2} compares the results of both approaches on the Line
Segment Detection problem. The figures show the probability of detection as a
function of the number of points within a candidate rectangle, and the number
of those points which are aligned. The size of the rectangle is cut at $60$
(this is the largest value for which we have samples along the whole vertical
axis), and of course the number of aligned points cannot exceed the size of the
rectangle.  The white region correspond to the detections in each case. For
example, both methods detect a segment when  there are at least 30 aligned
points in a rectangle with 40 points, and naturally there is a limit of 40
aligned points in that case. The 3rd figure is a comparison between the results
of both methods. It can be seen that again both criteria produce very similar
results and the a-contrario approach is a bit more sensitive, i.e.  produce
detections with a lower number of aligned points in the rectangle.

\section{When are MDL and a-contrario equivalent?}\label{sec:equivalence}

We have shown various examples where MDL and a-contrario lead to similar
formulations and very similar numerical results.  Nevertheless the formulations
are strictly different.  Indeed, MDL and a-contrario are in general different
theories.  In this section we present some conditions under which MDL and
a-contrario approaches result in exactly the same criterion.

As we observed before, MDL implies selecting among a family of possible
descriptions, the one with the shortest description of the \emph{whole} data.
On the other hand, the a-contrario approach concentrates on detecting
\emph{parts} of the data with anomalous statistics.  This is a key observation,
pointing to an important difference and also suggesting how to connect the two
approaches.  The connection can be obtained by forcing MDL to work also by
parts.  As we will see, given an a-contrario modeling, it is possible to build
a MDL modeling resulting in  exactly the same decision.\footnote{This is true
when handling discrete data.  The a-contrario approach can handle naturally
continue data, which imposes a quantization step in MDL.  In practice, this
difference is of little importance as most cases of interest can be stated in a
discrete way.}  The opposite is not true; MDL is a more general theory.

Following the description in \Cref{sec:nfa}, let $\mathbf{x}$ be a data vector
taking values in a finite alphabet $\mathcal{X}$ and let $\{\mathbf{x}_1,\,
\mathbf{x}_2,\ldots,\mathbf{x}_N\}$ be a family of $N$ parts.  The ordering
function $\xi$ acts on data vectors $\mathbf{x}_i\in\mathcal{X}^{n_i}$ and
produces real numbers $y_i = \xi(\mathbf{x}_i)$.  Finally, a stochastic model
$H_0$ is required for background data; here  we can follow Laplace's principle
of indifference and assume that all the elements in $\mathcal{X}^n$ are equally
probable under $H_0$.  A part $\mathbf{x}_i$ will be considered anomalous when
observing a value $y_i = \xi(\mathbf{x}_i)$ or larger is a rare event under
$H_0$.  To be precise, an anomaly is declared when
\begin{equation}\label{eq:nfa-condition}
    N \cdot \mathbb{P}[Y_i \geq y_i] \leq \varepsilon,
\end{equation}
where $Y_i$ is a random variable corresponding to $y_i$ under $H_0$.  Then, by
\Cref{prop:nfa}, the expected number of false alarms in $H_0$ is bounded by
$\varepsilon$.  As a consequence, $\varepsilon$ determines the mean number of
false alarms we are ready to accept per data set $\mathbf{x}$.

With these elements, we can now specify the equivalent MDL modeling. Three
assumptions are required:
\begin{enumerate}
    \item The vector $\mathbf{x}$ is coded as random data (the background),
          with the exception of some selected parts; those parts are selected
          from a family of $N$ and will be handled independently from each
          other.
    \item In each potential part $\mathbf{x}_i$, the configurations with
          larger $\xi(\mathbf{x}_i)$ should be favoured.
    \item In each part $\mathbf{x}_i$, the configurations to be coded
          differently than the background must use the same code-length $l_i$.
\end{enumerate}
Under these assumptions, to decide whether a given part should be described
differently than the background or not, both lengths must be compared.  When
treated as background, the code-length for the part is $\log
|\mathcal{X}^{n_i}|$.  On the other hand, when described as a particular part,
we need to specify which part is going to be described: this requires $\log N$
bits. Then, we need to specify the configuration of $\mathbf{x}_i$: this
requires $l_i$ bits.  Notice that the part needs not to be specified when coded
as background as all the elements of $\mathbf{x}$ not included in  a particular
part will be coded as background; only the parts to be coded differently must
be specified, and the rest is background. Thus, $\mathbf{x}_i$ will be
described as a particular part when
\begin{equation}\label{eq:codelength-mdl-as-nfa}
    \log N + l_i < \log|\mathcal{X}^{n_i}|.
\end{equation}
In other words, this makes sense when
$$
   l_i < \log|\mathcal{X}^{n_i}| - \log N,
$$
which means that at most $\frac{|\mathcal{X}^{n_i}|}{N}$ configurations of
$\mathbf{x}_i$ can be coded in that particular way.  Which configurations?  The
ones with the largest $\xi(\mathbf{x}_i)$ values.  When the following condition
$$
    \Big| \big\{ v \in \mathcal{X}^{n_i},\,\, \xi(v) \geq \xi(\mathbf{x}_i) \big\} \Big| <
    \frac{|\mathcal{X}^{n_i}|}{N}
$$
is satisfied, this implies that $\mathbf{x}_i$ is among the
$\frac{|\mathcal{X}^{n_i}|}{N}$ configuration with largest $\xi(\mathbf{x}_i)$
and should be coded as a especial part.  This in turn is equivalent to
$$
    N \frac{\Big| \big\{ v \in \mathcal{X}^{n_i},\,\, \xi(v) \geq \xi(\mathbf{x}_i) \big\} \Big|}{|\mathcal{X}^{n_i}|} < 1,
$$
which is equivalent to
$$
    N \cdot \mathbb{P}\big[\xi(\mathbf{X}_i) \geq \xi(\mathbf{x}_i)\big] < 1,
$$
where $\mathbf{X}_i$ is a random vector following $H_0$.  This condition is
equivalent to \cref{eq:nfa-condition} when setting $\varepsilon=1$ and the same
a-contrario criterion is obtained.

Notice that the same can be done when a non-uniform risk distribution $\eta_i$
is used.  Indeed, the condition $\sum_{i=1}^N \frac{1}{\eta_i} \leq 1$ makes
that $\log \eta_i$ satisfy the Kraft inequality and thus there is a prefix
coding for the parts with code-lengths $\log \eta_i$.  Using $\log \eta_i$
instead of $\log N$ in \cref{eq:codelength-mdl-as-nfa} leads to the condition
$$
    \eta_i \cdot \mathbb{P}\big[\xi(\mathbf{X}_i) \geq \xi(\mathbf{x}_i)\big] < 1,
$$
which again is the a-contrario criterion when setting $\varepsilon=1$.

From the MDL point of view, the a-contrario approach can be thought as a
particular strategy for modelling.  In the context of the MDL framework, this
strategy is often sub-optimal.  Nevertheless, the complexity of an optimal
modeling imposes very often in practice a sub-optimal approach in MDL
applications.  To prevent a combinatorial explosion, the modeling is often
 done by independent parts (as was the case, for example, in
\Cref{sec:mdl-lsd}).

The conditions required for the equivalence are not exactly satisfied in the
examples described in this work, only approximately in some of the cases.  As a
consequence, the criteria obtained are only similar but not equivalent.

\section{Discussion}\label{sec:discussion}

The MDL principle and the a-contrario approaches are very different theories,
in their philosophical and mathematical foundations.  Nevertheless, both share
similar characteristics.  Both can be used in model selection problems and in
detection problems.  Both require a modeling step and a given problem can be
modeled in various ways.  Tradition favors some kinds of models in MDL and
other kinds in a-contrario.  Here we made an effort to handle the same problems
by both approaches and enforce the modeling to be similar.  As expected, the
resulting criteria are different; surprisingly the resulting decisions are
however very similar.  This may be explained by Kolmogorov's definition of
randomness, as suggested in the introduction.  Indeed, there is a connection
between the compression resulting from a model and the non-accidentalness
evaluated by the same model.  Thus, a configuration considered as accidental in
the a-contrario approach will not lead to a shorter description than the random
model in MDL, leading to similar decisions.  Nevertheless, this conclusion is
only valid when the same (or similar) modeling is used; otherwise, a richer
family of models could handle more complex patterns, allowing for a shorter
description or identifying a non-accidental configuration.

Making the theoretically optimal decision according to both, MDL and
a-contrario usually require the evaluation of a very large number of events.
In practice, this is not possible and both criteria need to be applied to a
much smaller subset of events using heuristics to propose relevant candidates.
The real-life applications presented in this work fall into this category, as
the events involve the presence or absence of a large number of points or
segments in all possible configurations; a heuristic such as BSS or
FSS is thus needed.

In the MDL formulation, the detection or model selection is determined by the
modeling step (and the required heuristics in most cases).  The a-contrario
approach requires, in addition, setting the expected number of false alarms
threshold $\varepsilon$.  This is a clear advantage of the MDL approach as it
requires one less parameter to be set (except in some particular cases where
the ability to specify the false detection rate may be useful).  Nevertheless,
considering that the a-contrario formulation already includes the number of
tests performed, the reasonable range of values for $\varepsilon$ is quite
limited.  Moreover, given the usual logarithmic dependence on $\varepsilon$,
the actual impact of this parameter is very limited.  Indeed, the frontier
between a pattern that can be observed by accident or not is usually very
sharp.  This is confirmed in our experiments where $\varepsilon$ was always set
to one, and in all cases, MDL and a-contrario resulted in very similar
criteria.

A key limitation of MDL is in dealing with real-valued data.  As the
performance of a model is measured by how much it can compress the data, any
reasonable model has to be able to produce a code length that is significantly
shorter than the ``raw'' description of the data.  For example, if we have a
one megapixel 8-bit image, any description over 8 megabits should be discarded.
This is already a problem with MDL on signals over large alphabets but becomes
virtually impossible if the data is real-valued, simply because describing real
numbers requires infinite precision.  Quantization is therefore a mandatory
step, and optimum quantization is still an open problem in Information Theory.
On the other hand, real numbers are not a problem in the a-contrario framework
as long as the events to be detected can be cast as thresholds on appropriate
random variables.

Several works discuss the connection between the parsimony principle and the
likelihood principle~\cite{grunwald07,chater,van-der-Helm2020}.  Similarly, the
present work draws the connections between the parsimony principle and the
non-accidentalness principle as expressed in the MDL and a-contrario
formalisms.

A key difference is that the MDL principle focuses on the best interpretation
for the whole data while the a-contrario approach concentrates on parts of the
data with anomalous statistics.  To prevent a combinatorial explosion, in
practice MDL is often dictated to evaluate the data by parts.  Again, the
theoretical differences vanish in real applications.  The examples presented in
this work and the conditions of equivalence in \Cref{sec:equivalence} suggest
that when working by parts, both MDL and a-contrario share a common ground.
More generally, any departure from randomness should allow for a shorter
description in MDL using an appropriate modelling; the same departure from
randomness should be detected as an anomaly by an appropriate a-contrario
modelling.  In a sense, both approaches embrace a common rationale which may be
described with the words of Dennett: ``Any nonrandomness in the flux [of energy
striking one's sensory organs] is a real pattern that is \emph{potentially
useful} information for some possible creature or agent to exploit in
anticipating the future.'' \cite[p.128]{dennett2017}

\section{Conclusion}\label{sec:conclusion}

MDL and a-contrario  are seemingly very different approaches, typically used in
different scenarios. The former try to codify the whole data and the later
concentrates on the detection of some anomalous structures. After having
compared both criteria, under different settings, both analytically and in
practice, we have found that they are, in fact, closely related. Our initial
discussion makes it clear that both methods share a common root in the
Algorithmic Complexity theory. An in-depth analysis of both methods in toy
examples has shown that there are significant connections in the tools and the
mathematical formulations behind the metrics used (code-length and NFA).  When
applied, we found that both methods yield similar results in all the scenarios
studied; in some cases, the results are almost identical.  Last but not least,
we have also shown that both MDL and a-contrario  methodologies can be used
interchangeably for detection  (single hypothesis) and model selection
(multiple hypotheses) scenarios, without significant theoretical or practical
complications.

Having established these connections opens up new and exciting lines of work
involving the cross-dissemination of these methods to new domains. Examples of
these are the application of MDL tools for Computer Vision problems or using
the a-contrario formalism for tackling a wide range of model selection problems
in and outside the fields of Computer Vision and Image Processing.

\appendix

\section{Stirling's approximation for binomial terms}\label{sec:stirling}

Here we develop the Stirling approximation of ${n \choose k}$.  First, we have
the basic Stirling approximation for $n!$:
$$
   n! \approx \sqrt{2\pi n} e^{-n} n^n.
$$
When plugged into ${n \choose k} = \frac{n!}{k!(n-k)!}$ we obtain:
\begin{align*}
\frac{n!}{k!(n-k)!} &\approx \frac{\sqrt{2\pi n} e^{-n} n^n }{\sqrt{2\pi k} e^{-k} k^k  \sqrt{2\pi (n-k)} e^{-(n-k)} (n-k)^{n-k} } \\
&= \frac{1}{\sqrt{2\pi}} \sqrt{\frac{n}{k(n-k)}} \frac{ n^n }{k^k (n-k)^{n-k} } \\
&= \frac{1}{\sqrt{2\pi}} \sqrt{\frac{n}{k(n-k)}} \left(\frac{n}{k}\right)^k\left(\frac{n}{n-k}\right)^{n-k}\\
\end{align*}
Taking logarithms, we arrive at:
$$
   \log {n \choose k} \approx \frac{1}{2}\log \frac{1}{2\pi} +
\frac{1}{2}\log \frac{n}{k(n-k)} +k \log \frac{n}{k} + (n-k)\log \frac{n}{n-k}.
$$

\section{Bounds on g(k,n)}\label{sec:g-bounds}

Recall that $g(k,n)=\log\frac{n^3}{k(n-k)}$, which is a convex function of
$(k,n)$. Thus, the terms $g(k_0,n_0) + g(k_1,n_1)$ form a convex function on
$(k_0,n_0,k_1,n_1)$ which is block-symmetric in $(k_0,n_0)$ and $(k_1,n_1)$ as
both are interchangeable.  Since the restrictions in place establish that
$k_0+k_1=k$ and $n_0+n_1=n$, the minimum is attained at the mid-point
$(k_0,n_0)=(k_1,n_1)=(k/2,n/2)$.  Thus, for fixed $k$,$n$, we have
$g(k/2,n/2)=\log\frac{n^3/2}{k(n-k)}=g(k,n)-1$.  Thus,
$$
   \begin{array}{cl}
     \min_{k_0,n_0,k_1,n_1} & g(k_0,n_0) + g(k_1,n_1) \\
     s.t. & k_0+k_1 = k, n_0+n_1 = k
   \end{array} = 2g(k,n) - 2
$$
and
$$
\frac{1}{2}\Big[g(k_0,n_0) + g(k_1,n_1) - g(k,n)\Big]
\geq
\frac{1}{2}[2g(k,n) - 2 - g(k,n)]
= \frac{1}{2} g(k,n) - 1.
$$
Now, $g(k,n)$ is minimized when $k=n/2$ with $g(\frac{n}{2},n) = 2 + \log n$.
Thus we have a lower bound (which is tight when $n$ is a multiple of $4$):
\begin{equation}\label{eq:mdl-devil-term-2}
\frac{1}{2}\Big[g(k_0,n_0) + g(k_1,n_1) - g(k,n)\Big]
\geq
\frac{1}{2} \log n.
\end{equation}
The upper bound can be worked in a similar way.  Since, for fixed $n$, $g(k,n)$
is symmetric and strictly convex around $k=n/2$, we know that the maxima occur
at the extreme points of the feasible set, $1 \leq k \leq n-1$.  Thus, $\max_k
g(k,n) = g(1,n) = g(n-1,n) = \log \frac{n^3}{n-1}$.  Applying analogous
arguments as before, the maxima of the symmetric function $g(k_0,n_0) +
g(k_1,n_1)$ must occur when $n_1 = n_0=n/2$ and $k_0=k_1=1$ so that
$$
  \begin{array}{cl}
    \max_{k_0,n_0,k_1,n_1} & g(k_0,n_0) + g(k_1,n_1)   \\
     s.t. & k_0+k_1 = k,\,\, n_0+n_1 = n
  \end{array} = 2 \log \frac{n^3}{4(n-2)}.
$$
Recalling that $g(k,n)$ is minimized by $g(\frac{n}{2},n) = 2 + \log n$, we can
now maximize the whole term over $2 \leq k \leq n-2$ and obtain its upper
bound:
$$
  \frac{1}{2}\Big[g(k_0,n_0) + g(k_1,n_1) - g(k,n)\Big]
  \leq
  \frac{2}{2} \log \frac{n^3}{4(n-2)} - \frac{1}{2} (2 + \log n)
  = \log \frac{n^{\frac{5}{2}}}{4(n-2)} - 1.
$$
The remaining MDL term \cref{eq:mdl-devil-term-1} has now been bounded as
follows:
\begin{equation}
    \log n
    \leq
    \frac{1}{2}\Big[g(k_0,n_0) + g(k_1,n_1) - g(k,n)\Big]
    \leq
    \log \frac{n^{\frac{5}{2}}}{4(n-2)} - 1
    \approx
    \frac{3}{2}\log n - 3.
\end{equation}

\bibliographystyle{siamplain}

\begin{thebibliography}{10}

\bibitem{albert-hoffman1995}
{\sc M.~K. Albert and D.~D. Hoffman}, {\em Genericity in spatial vision}, in
  Geometric Representations of Perceptual Phenomena: Articles in Honor of Tarow
  Indow's 70th Birthday, D.~Luce, K.~Romney, D.~Hoffman, and {D'Z}mura M.,
  eds., Erlbaum, 1995, pp.~95--112.

\bibitem{barron98}
{\sc A.~Barron, J.~Rissanen, and B.~Yu}, {\em The minimum description length
  principle in coding and modeling}, IEEE Trans. IT, 44 (1998), pp.~2743--2760.

\bibitem{chaitin1969}
{\sc G.~J. Chaitin}, {\em On the simplicity and speed of programs for computing
  infinite sets of natural numbers}, Journal of the ACM, 16 (1969),
  pp.~407--422.

\bibitem{chaitin}
{\sc G.~J. Chaitin}, {\em Algorithmic information theory}, {IBM} Journal of
  Research and Development, 21 (1977), pp.~350--359.

\bibitem{chater}
{\sc N.~Chater}, {\em Reconciling simplicity and likelihood principles in
  perceptual organization}, Psychological Review, 103 (1996), pp.~566--581.

\bibitem{cover06}
{\sc T.~Cover and J.~Thomas}, {\em Elements of information theory}, John Wiley
  and Sons, Inc., 2~ed., 2006.

\bibitem{cover73}
{\sc T.~M. Cover}, {\em Enumerative source encoding}, IEEE Trans. IT, 19
  (1973), pp.~73--77.

\bibitem{cover91}
{\sc T.~M. Cover and J.~A. Thomas}, {\em Elements of information theory}, John
  Wiley and Sons, Inc., 1991,
  \url{/bib/private/cover/Wiley.Interscience.Elements.of.Information.Theory.Jul.2006.eBook-DDU.pdf}.

\bibitem{dennett2017}
{\sc D.~C. Dennett}, {\em From Bacteria to Bach and Back}, W. W. Norton \&
  Company, 2017.

\bibitem{DMM2000}
{\sc A.~Desolneux, L.~Moisan, and J.-M. Morel}, {\em Meaningful alignments},
  International Journal of Computer Vision, 40 (2000), pp.~7--23.

\bibitem{DMM_book}
{\sc A.~Desolneux, L.~Moisan, and J.-M. Morel}, {\em From {G}estalt Theory to
  Image Analysis, a Probabilistic Approach}, Springer, 2008.

\bibitem{Devernay95anon-maxima}
{\sc F.~Devernay}, {\em A non-maxima suppression method for edge detection with
  sub-pixel accuracy}, tech. report, INRIA Research Rep. 2724, Sophia
  Antipolis, 1995.

\bibitem{gordon2007}
{\sc A.~Gordon, G.~Glazko, X.~Qiu, and A.~Yakovlev}, {\em Control of the mean
  number of false discoveries, {B}onferroni and stability of multiple testing},
  The Annals of Applied Statistics, 1 (2007), pp.~179--190.

\bibitem{lsd-book}
{\sc R.~Grompone~von Gioi}, {\em A contrario line segment detection}, Springer,
  2014.

\bibitem{lsd-ipol}
{\sc R.~Grompone~von Gioi, J.~Jakubowicz, J.-M. Morel, and G.~Randall}, {\em
  {LSD}: a {L}ine {S}egment {D}etector}, Image Processing On Line,  (2012),
  \url{http://dx.doi.org/10.5201/ipol.2012.gjmr-lsd}.

\bibitem{devernay-ipol}
{\sc R.~Grompone~von Gioi and G.~Randall}, {\em {A Sub-Pixel Edge Detector: an
  Implementation of the Canny/Devernay Algorithm}}, {Image Processing On Line},
  7 (2017), pp.~347--372, \url{https://doi.org/10.5201/ipol.2017.216}.

\bibitem{grosjean-moisan}
{\sc B.~Grosjean and L.~Lionel~Moisan}, {\em A-contrario detectability of spots
  in textured backgrounds}, Journal of Mathematical Imaging and Vision, 33
  (2009), \url{https://doi.org/10.1007/s10851-008-0111-4}.

\bibitem{grunwald07}
{\sc P.~Gr\"{u}nwald}, {\em The Minimum Description Length Principle}, {MIT}
  Press, June 2007.

\bibitem{hastie08}
{\sc T.~Hastie, R.~Tibshirani, and J.~Friedman}, {\em The Elements of
  Statistical Learning: Data Mining, Inference and Prediction}, Springer,
  2~ed., Feb. 2009.

\bibitem{HT87}
{\sc Y.~Hochberg and A.~C. Tamhane}, {\em Multiple comparison procedures}, John
  Wiley \& Sons, New York, 1987.

\bibitem{Hoeffding}
{\sc W.~Hoeffding}, {\em {Probability Inequalities for Sums of Bounded Random
  Variables}}, Journal of the American Statistical Association, 58 (1963),
  pp.~13--30, \url{https://doi.org/10.1080/01621459.1963.10500830},
  \url{https://www.tandfonline.com/doi/abs/10.1080/01621459.1963.10500830}.

\bibitem{kolmogorov}
{\sc A.~N. Kolmogorov}, {\em Three approaches to the definition of the concept
  ``quantity of information''}, Probl. Peredachi Inf., 1 (1965), pp.~3--11.

\bibitem{lsd}
{\sc J.~Lezama, J.~M. Morel, G.~Randall, and R.~Grompone~von Gioi}, {\em A
  contrario 2d point alignment detection}, Pattern Analysis and Machine
  Intelligence, IEEE Transactions on, 37 (2015), pp.~499--512.

\bibitem{li-vitanyi}
{\sc M.~Li and P.~Vit\'anyi}, {\em An Introduction to {K}olmogorov Complexity
  and Its Applications}, Springer, 4th~ed., 2019.

\bibitem{Lowe85}
{\sc D.~Lowe}, {\em Perceptual Organization and Visual Recognition}, Kluwer
  Academic Publishers, 1985.

\bibitem{rissanen78}
{\sc J.~Rissanen}, {\em Modeling by shortest data description}, Automatica, 14
  (1978), pp.~465--471.

\bibitem{rissanen84}
{\sc J.~Rissanen}, {\em Universal coding, information, prediction, and
  estimation}, IEEE Trans. IT, 30 (1984), pp.~629--636.

\bibitem{rissanen86}
{\sc J.~Rissanen}, {\em Stochastic complexity in modeling}, Annals of
  Statistics, 14 (1986), pp.~1080--1100.

\bibitem{rissanen92}
{\sc J.~Rissanen}, {\em Stochastic complexity in statistical inquiry},
  Singapore: World Scientific, 1992.

\bibitem{solomonoff1}
{\sc R.~Solomonoff}, {\em A formal theory of inductive inference. part i},
  Information and Control, 7 (1964), pp.~1 -- 22,
  \url{https://doi.org/https://doi.org/10.1016/S0019-9958(64)90223-2},
  \url{http://www.sciencedirect.com/science/article/pii/S0019995864902232}.

\bibitem{solomonoff2}
{\sc R.~Solomonoff}, {\em A formal theory of inductive inference. part ii},
  Information and Control, 7 (1964), pp.~224 -- 254,
  \url{https://doi.org/https://doi.org/10.1016/S0019-9958(64)90131-7},
  \url{http://www.sciencedirect.com/science/article/pii/S0019995864901317}.

\bibitem{van-der-Helm2020}
{\sc P.~A. van~der Helm}, {\em Dubious claims about simplicity and likelihood:
  Comment on pinna and conti (2019)}, Brain Sciences, 10 (2020).
\newblock https://doi.org/10.3390/brainsci10010050.

\bibitem{wagemans-nonaccidental}
{\sc J.~Wagemans}, {\em Perceptual use of nonaccidental properties}, Canadian
  Journal of Psychology, 46 (1992), pp.~236--279.

\bibitem{non-computable}
{\sc A.~Zvonkin and L.~Levin}, {\em The complexity of finite objects and the
  development of the concepts of information and randomness by means of the
  theory of algorithms}, Russian Math. Surveys, 25 (1970), pp.~83--124.

\end{thebibliography}

\end{document}